\documentclass[10pt, journal, compsoc]{IEEEtran}
\usepackage{amsmath,amsfonts}
\usepackage{algorithmic}
\usepackage{algorithm}
\usepackage{array}
\usepackage[caption=false,font=normalsize,labelfont=sf,textfont=sf]{subfig}
\usepackage{textcomp}
\usepackage{stfloats}
\usepackage{url}
\usepackage{verbatim}
\usepackage{graphicx}
\usepackage{cite}
\usepackage{amsmath,amssymb,amsfonts}
\usepackage{multirow}
\usepackage{booktabs}
\usepackage[table]{xcolor}
\usepackage{comment}
\usepackage{longtable}
\usepackage{threeparttable}
\usepackage{array}
\usepackage{hyperref}
\hypersetup{hidelinks}
\usepackage[switch, mathlines]{lineno}
\usepackage[numbers]{natbib}
\usepackage{enumitem}

\providecommand{\internallinenumbers}{}

\begin{document}

\title{Large Language Models as a Universal Searcher for
Open-World Automated Machine Learning}

\author{Yuxuan~Hu,
        Jihao~Liu,
        Weikang~Shi,
        Xijie~Huang,
        Linda~Wei,
        Mingjie~Zhan,
        Aojun~Zhou,
        Rui~Liu,
        and~Hongsheng~Li,~\IEEEmembership{Member,~IEEE}%
\thanks{This work has been submitted to the IEEE for possible publication. Copyright may be transferred without notice, after which this version may no longer be accessible.}
\thanks{This work is partially supported by the National Key R\&D Program of China (No. 2025ZD0122902), in part by NSFC-RGC Project N\_CUHK498/24, in part by Guangdong Basic and Applied Basic Research Foundation (No. 2023B1515130008, XW), in part by Shenzhen Loop Area Institute, and in part by the Centre for Perceptual and Interactive Intelligence (CPII) Ltd under the Innovation and Technology Commission (ITC)'s InnoHK.}%
\thanks{Yuxuan Hu, Jihao Liu, Weikang Shi, Xijie Huang, Linda Wei, Mingjie Zhan, Aojun Zhou, and Rui Liu are with the Multimedia Laboratory (CUHK MMLab), The Chinese University of Hong Kong, Shatin, Hong Kong (e-mail: aojunzhou@gmail.com; ruiliu011@gmail.com).}%
\thanks{Hongsheng Li is with the Multimedia Laboratory (CUHK MMLab), The Chinese University of Hong Kong, Shatin, Hong Kong, and also with the Shenzhen Loop Area Institute, Shenzhen, China (e-mail: hsli@ee.cuhk.edu.hk).}%
}


\maketitle

\begin{abstract}
Developing effective machine learning models requires complex manual tuning, while current Automated Machine Learning (AutoML) frameworks remain domain-specific and isolate model architecture, training hyperparameter, and data ratio optimization. Here, we bridge this gap by proposing a universal, large language model (LLM)-based AutoML framework that jointly optimizes these configurations for the first time without the need for task-specific fine-tuning. We propose NCode, a unified numerical encoding scheme that translates diverse AutoML design choices into universal discrete sequences and effectively reframes the optimization process as a unified ranking task. Trained solely on optimization trajectories derived from a limited, single-domain vision task, the proposed LLM-based searcher demonstrates impressive cross-task, cross-modality and joint search-space optimization generalizability, achieving state-of-the-art performance across image classification, image segmentation, speaker verification and text-to-image benchmarks. Crucially, it exhibits superior scalability in the large-scale training and joint optimization of a billion-parameter Diffusion Transformer, requiring 896 GPU-days and attaining a state-of-the-art GenEval score of 0.91. By leveraging reinforcement learning to instill Chain-of-Thought reasoning, we provide a systematic way to interpret the LLM-based searcher's internal decision-making mechanisms, revealing its optimization strategy shifted from the early-stage diversity-driven exploration to the late-stage anchor exploitation. Our proposed framework provides a scalable, generalizable, and interpretable pathway toward general-purpose open-world machine learning optimization.
\end{abstract}

\begin{IEEEkeywords}
Automated machine learning, open-world generalization, joint optimization, large language models.
\end{IEEEkeywords}

\section{Introduction}
\IEEEPARstart{R}{ecent} advancements in Machine Learning (ML) have revolutionized diverse domains, including computer vision, natural language processing, and audio analysis. Despite the progress, developing effective ML models remains a complex endeavor, often requiring extensive human experience in designing network architectures, tuning hyperparameters, and organizing training data. Consequently, Automated Machine Learning (AutoML) has emerged as a pivotal solution, aiming to automate such design choices of ML pipelines with minimal human intervention. 

While intensive explorations in AutoML have been made, particularly in Neural Architecture Search (NAS) for architecture design~\cite{zoph2017neural,zoph2018learning,liu2018darts} and Hyperparameter Optimization (HPO) for determining training hyperparameters~\cite{chen2022towards}, the field still lacks a general-purpose AutoML framework. This gap has become increasingly challenging in the era of foundation models. Currently, the realization of such an open-domain system is impeded by three fundamental obstacles:

\begin{itemize}
    \item \textbf{Open-world problem generalization.} Existing AutoML approaches are typically optimized for specific search spaces, lacking the flexibility to generalize across diverse tasks, datasets and architectures. For instance, the popular gradient-based NAS method DARTS~\cite{liu2018darts} achieves 97.24\% accuracy on the CIFAR-10 test set within its original search space, yet suffers a severe performance collapse to 39.77\% when evaluated on the widely used NAS-Bench-201 benchmark~\cite{dong2020bench}. Even though both settings share identical dataset and optimizers, this sharp decline underscores the inherent fragility of current NAS and AutoML techniques when subjected to structural shifts in the search space. Consequently, this limitation hinders the application of AutoML methods in open-world problems that data distributions, architecture search spaces and task definitions are novel or unseen. 
    
    \item \textbf{Intricate interdependence of model optimization choices.} The optimal performance of a model is the result of the complex interplay between model architectures, training hyperparameters, and data ratios, etc. Such highly non-linear interactions are difficult to optimize in isolation. Existing AutoML methods fail to capture the synergies and often lead to sub-optimal local minima when components are optimized separately.
    
    \item \textbf{The prohibitive cost of AutoML training data acquisition.} Creating AutoML training datasets is computationally expensive and time-consuming, as it involves the repetitive process of model training and performance evaluation for every new task. The high cost hinders the development and scalability of learning-based AutoML methods.
\end{itemize}

To overcome these barriers, the field is witnessing a paradigm shift from specialized optimization algorithms toward more generalizable approaches. In particular, Large Language Models (LLMs) were leveraged to optimize AutoML problems to take advantage of their strong reasoning capabilities~\cite{zhang2023automl,li2026collm}. However, they remain heavily dependent on task-specific training, fine-tuning, and manual prompt engineering. As a result, there has been little deployment of AutoML frameworks that can navigate open-world problems spanning across multiple architectures, tasks, and modalities.

\begin{figure*}[!htbp]
    \centering
    \includegraphics[width=0.9\textwidth]{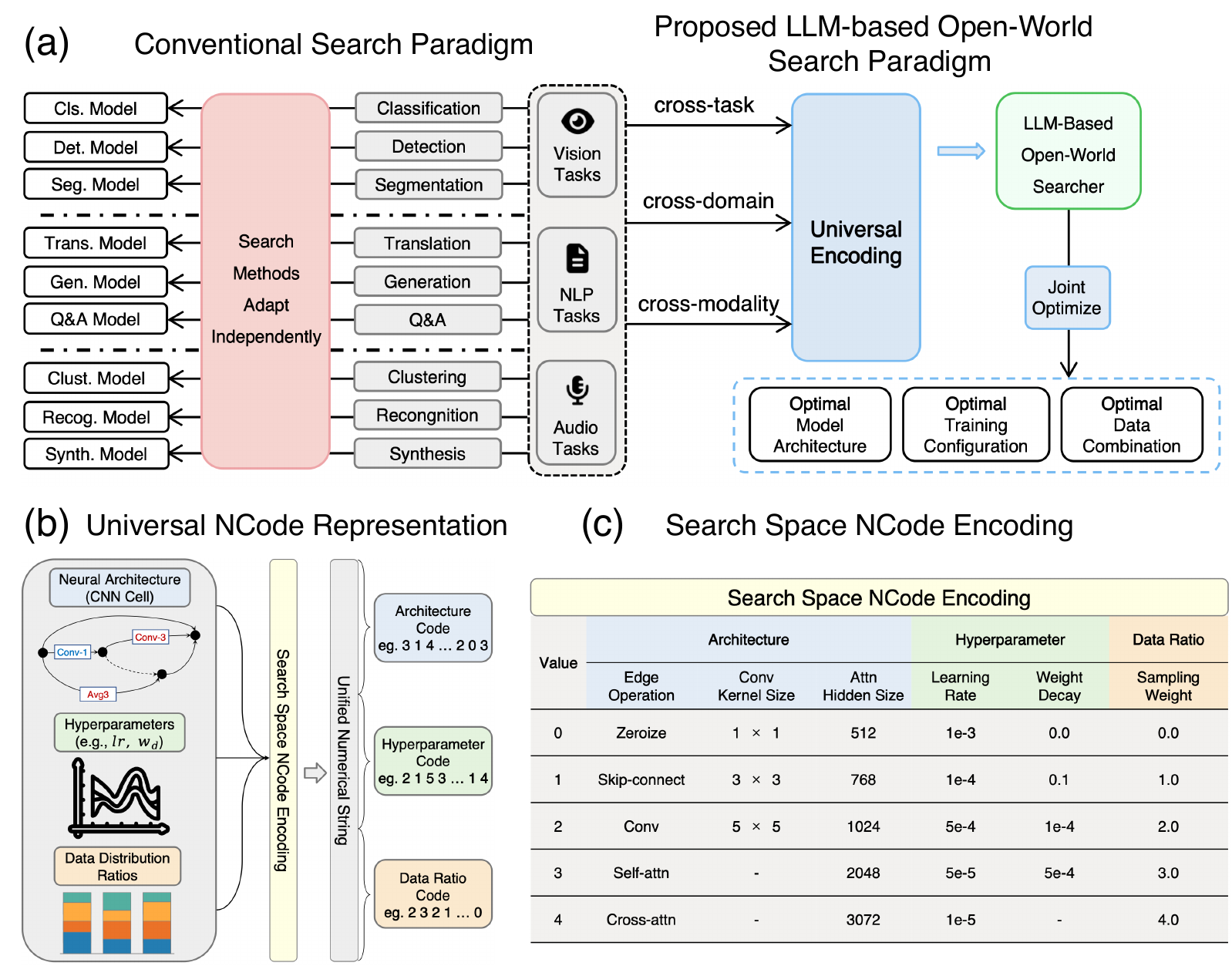} 
    \caption[]{Overview of the proposed LLM-based open-world AutoML framework and the universal NCode representation. (a) Conventional AutoML paradigms typically require separate adaptation of the model for different tasks. In contrast, our proposed open-world LLM-based search paradigm uses universal encoding to jointly optimize model architectures, training configurations, and data combinations across tasks, domains, and modalities. (b) NCode maps heterogeneous AutoML design choices, including architectures, hyperparameters, and data sampling weights, into unified discrete numerical sequences. (c) Illustrative encoding process for models with varying topologies, operations, training hyperparameters, and data ratios. Note that the depicted codebook is a simplified excerpt; the functional codebook used in our experiments is more extensive.}
    \label{fig:overall_overview}
\end{figure*}

In this work, we introduce an open-world LLM-based AutoML framework that is capable of jointly optimizing the complex combination of model architectures, training hyperparameters, data ratios for different tasks without task-specific fine-tuning for the first time. To enable the optimization over the multiple intrinsically diverse search spaces, we introduce NCode, an encoding scheme that maps heterogeneous design choices to their respective indices, representing them as unified numerical sequences (Fig.~\ref{fig:overall_overview}b,c). By converting diverse AutoML tasks into a universal discrete optimization problem, NCode allows LLMs to easily generalize to unseen open-world domains without domain-specific fine-tuning (Fig.~\ref{fig:overall_overview}a). To train the LLM-based searcher, we curate optimization trajectories from the NAS-Bench-301 dataset~\cite{zela2020surrogate}, which provides model architectures and their corresponding image classification performances on the CIFAR-10 dataset. We model the model arch instances as a densely connected Directed Acyclic Graph (DAG) and apply stochastic edge pruning to generate diverse, localized architecture search spaces for rich trajectory formulation. Furthermore, for the LLM-based searcher's outputs, we reframe the discrete NCode optimization as a ranking problem, where the LLM identifies superior candidates from a candidate pool rather than directly predicting their NCodes, a setting demonstrated to be significantly simpler to learn than directly generating full NCodes. Consequently, despite being trained exclusively on single-domain vision data, our LLM searcher generalizes effectively to cross-domain joint optimization tasks where data collection is typically cost-prohibitive.

A preliminary version of our system that focused only on model architecture search was presented in EMNLP2025~\cite{hu2025lm}. The current study unveils the unified LLM-based searcher's capabilities in significantly more complex, heterogeneous search spaces and mastering the synergy between model architectures, training hyperparameters, and data ratios across seven domains, four tasks and three modalities (vision, text, audio) without fine-tuning. More importantly, our LLM-based searcher is trained to explain its optimization rationales by learning explainable Chain-of-Thought (CoT) via Group Relative Policy Optimization (GRPO) ~\cite{shao2024deepseekmath}. Our analysis revealed how the model navigates the solution space for optimization and offers critical insights. Extensive experiments provide compelling evidence and demonstrate that our framework achieves state-of-the-art performance across a diverse suite of open-world machine learning problems.

We highlight our major contributions as follows:

\begin{enumerate}
    \item Unified Problem Representation: We introduce a universal numerical encoding that maps model architectures, hyperparameters, and data ratios into a unified discrete space, allowing the AutoML framework to jointly optimize diverse design choices across distinct search spaces.

    \item Search Trajectory Curation: We introduce a pruning-based architecture sampling strategy to curate rich, diverse optimization trajectories, mitigating the prohibitive cost of AutoML data collection. By simultaneously reframing the LLM's objective from direct sequence generation to conditional ranking, we boost the searcher's capacity to generalize across complex optimization landscapes.

    \item Extensive Empirical Validation: By deploying our approach across real-world machine learning tasks in diverse scenarios, and achieving state-of-the-art (SOTA) performances, we demonstrate the framework's robust zero-shot generalizability. 

    \item Interpretability via RL-driven Chain-of-Thought: To unveil the internal decision-making mechanism of the LLM-based searcher, we leverage Group Relative Policy Optimization (GRPO) to instill Chain-of-Thought (CoT) reasoning without requiring human annotation. This self-explaining capability provides critical insights into the model's learned optimization strategies, specifically revealing a dynamic shift from early-stage diversity-driven exploration to late-stage anchor exploitation, and highlights the inherent limitations of conventional discrete optimization methods in existing AutoML frameworks.
\end{enumerate}

\section{Related Work}
\subsection{Traditional Automated Machine Learning}
Automated Machine Learning (AutoML) has evolved significantly, progressing from early reinforcement learning~\cite{zoph2017neural,zela2018towards} and gradient-based Neural Architecture Search (NAS)~\cite{liu2018darts} to Transformer-based Hyperparameter Optimization (HPO)~\cite{chen2022towards}. More recently, the field has increasingly embraced advanced generative paradigms, such as leveraging multi-conditioned graph diffusion models~\cite{asthana2024multi} to sample candidate neural structures. Despite these advancements, conventional AutoML approaches remain inherently tailored to predefined, task-specific search spaces. Furthermore, even contemporary generative NAS models typically treat architecture search and training configuration tuning as disjoint problems. This isolation fails to capture the intricate synergies among heterogeneous components, restricting their applicability in open-world scenarios involving unseen modalities or structural shifts. In contrast to optimizing components in silos, our framework introduces NCode to unify these heterogeneous design choices into a shared numerical space, enabling holistic joint optimization.

\subsection{LLM-Based Automated Machine Learning.}
Recent years have witnessed a growing interest in leveraging Large Language Models (LLMs) for Neural Architecture Search (NAS) and Automated Machine Learning (AutoML), with representative frameworks including AutoML-GPT~\cite{zhang2023automl}, LLMatic~\cite{nasir2024llmatic}, and CoLLM-NAS~\cite{li2026collm}. Despite these advances, existing LLM-based searchers suffer from key limitations: they either rely heavily on task-specific fine-tuning or formulate search as a direct textual generation task, which is prone to producing invalid configurations and exhibiting high learning complexity in vast discrete spaces. While our preliminary work, LM-Searcher~\cite{hu2025lm}, initiated the exploration of cross-domain NAS via unified numerical encoding, it remained restricted to architecture-only search. To overcome these bottlenecks, the proposed framework unifies heterogeneous configurations into a discrete NCode space and reframes the search as a conditional candidate ranking problem. This design achieves robust zero-shot joint optimization of architectures, hyperparameters, and data ratios across diverse domains and modalities without task-specific adaptation.

\subsection{Explainability and Chain-of-Thought Reasoning.}
Traditional AutoML algorithms and neural search agents are frequently criticized for their black-box nature, providing minimal intrinsic rationale for their optimization trajectories~\cite{hutter2014efficient}. While Chain-of-Thought (CoT) prompting~\cite{wei2022chain} and reinforcement learning algorithms like Group Relative Policy Optimization (GRPO)~\cite{shao2024deepseekmath} have boosted logical reasoning in LLMs, applying self-explaining CoT to interpret AutoML search logic remains under-explored. To bridge this gap, the proposed framework leverages GRPO to post-train the LLM-based searcher to articulate its reasoning steps in natural language. By taxonomizing the resulting CoT outputs into distinct behavioral patterns, we provide interpretability into LLM-driven optimization mechanisms.

\begin{figure*}[!htbp]
    \centering
    \includegraphics[width=0.9\textwidth]{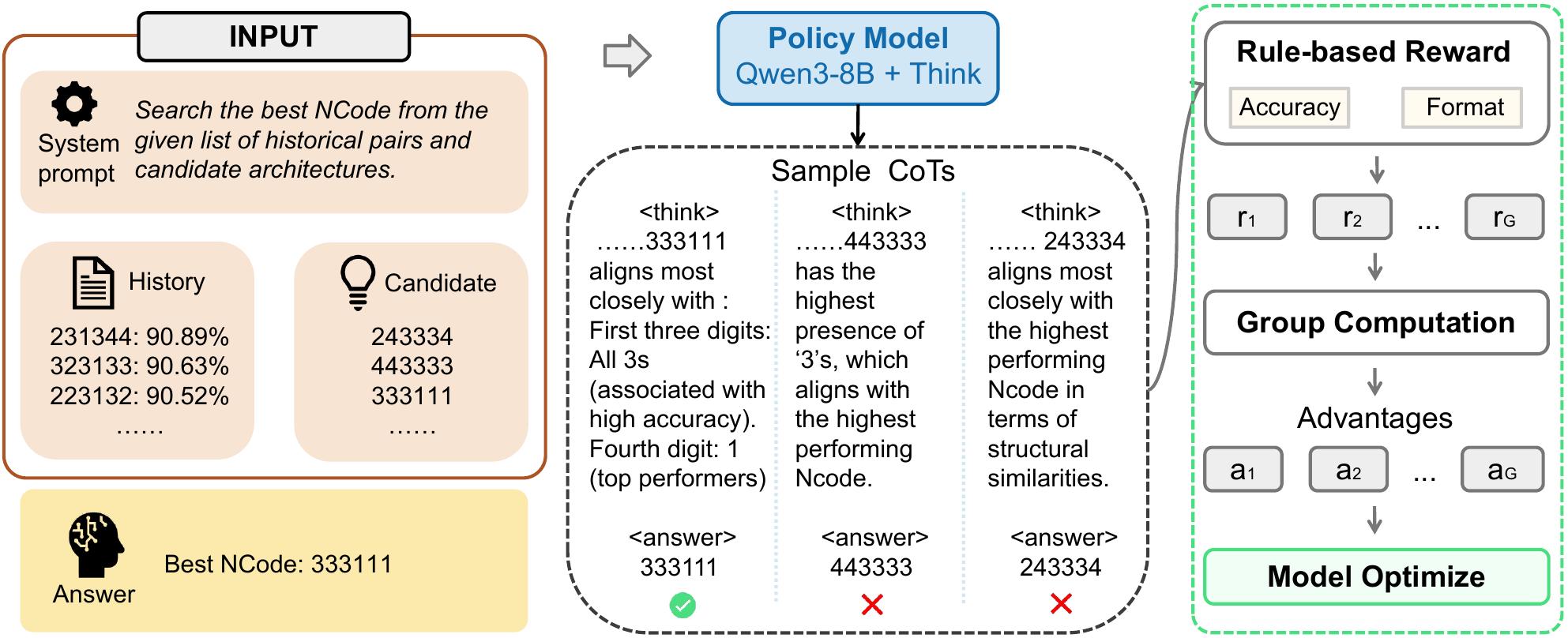} 
    \caption[]{Overview of post-training the LLM with CoT reasoning capabilities using Group Relative Policy Optimization. A Qwen3-8B model with native thinking abilities is fine-tuned on our curated dataset via reinforcement learning to articulate its reasoning paths. Given the system prompt, historical context, and candidate configurations represented as NCodes, the policy model generates multiple candidate responses. The total reward is calculated as a weighted sum of the accuracy reward (whether the model selects the optimal configuration from the candidate pool) and the format reward (whether the model outputs a complete reasoning process and the final answer in the correct format).}
    \label{fig:interpretability_pipeline}
\end{figure*}

\section{The LLM-based Searcher framework}
\subsection{Overview}
\label{sec:overview}
To enable the joint optimization of complex AutoML design choices including model architectures, training hyperparameters, and data ratios, we propose a universal, LLM-based open-world search framework. The core of our framework is a unified numerical encoding scheme termed NCode, which maps heterogeneous configurations into discrete sequences. During the search phase, the LLM-based searcher iteratively analyzes a historical buffer of evaluated NCode-reward pairs to autonomously identify and select the most promising configuration from an unevaluated candidate pool. Rather than relying on domain-specific finetuning, our LLM-based searcher is trained to implicitly learn optimization trajectories and reasoning strategies, allowing it to seamlessly generalize to unseen configurations and modalities.

\subsection{Universal NCode Representation}
\label{sec:ncode_representation}
To unify the representation of heterogeneous AutoML design choices (e.g., model architecture, training hyperparameters and data ratios) and task domains, we uniformly encode all configurations into discrete numerical sequences termed NCode. Formally, let a given AutoML search space be denoted as $\mathcal{S}$. An AutoML configuration $a \in \mathcal{S}$ can be defined by a set of $n$ decisions, which may represent operational primitives, network connectivity, training module-specific hyperparameters or data ratios. We define an encoding function $f: \mathcal{S} \rightarrow \mathcal{C}$, where $\mathcal{C}$ is the universal space of NCode sequences. The function maps the configuration to a string of discrete indices, denoted as $c = (c_1, c_2, \dots, c_n)$, where each element $c_i$ is a numerical index corresponding to a selected configuration option of the $i$-th configuration. 

NCode enables the joint states of model topologies, operations, training hyperparameters, and data ratios. To represent network topologies, networks can be modeled as Directed Acyclic Graphs (DAGs), where each edge represents an operation and each node represents a summed feature map. For instance (Fig.~\ref{fig:overall_overview}b), in the NAS-Bench-201 search space, each architecture is defined by assigning one of five distinct operations to each of the six edges within a cell. The NCode representation establishes a mapping between these operations and a set of integers. Notably, by framing the search as a black-box combinatorial optimization problem, NCode natively extends to jointly encode other heterogeneous configurations. Fig.~\ref{fig:overall_overview}c illustrates how NCode jointly encodes model architectures, training hyperparameters, and data ratios.

\begin{figure*}[!htbp]
    \centering
    \includegraphics[width=0.9\textwidth]{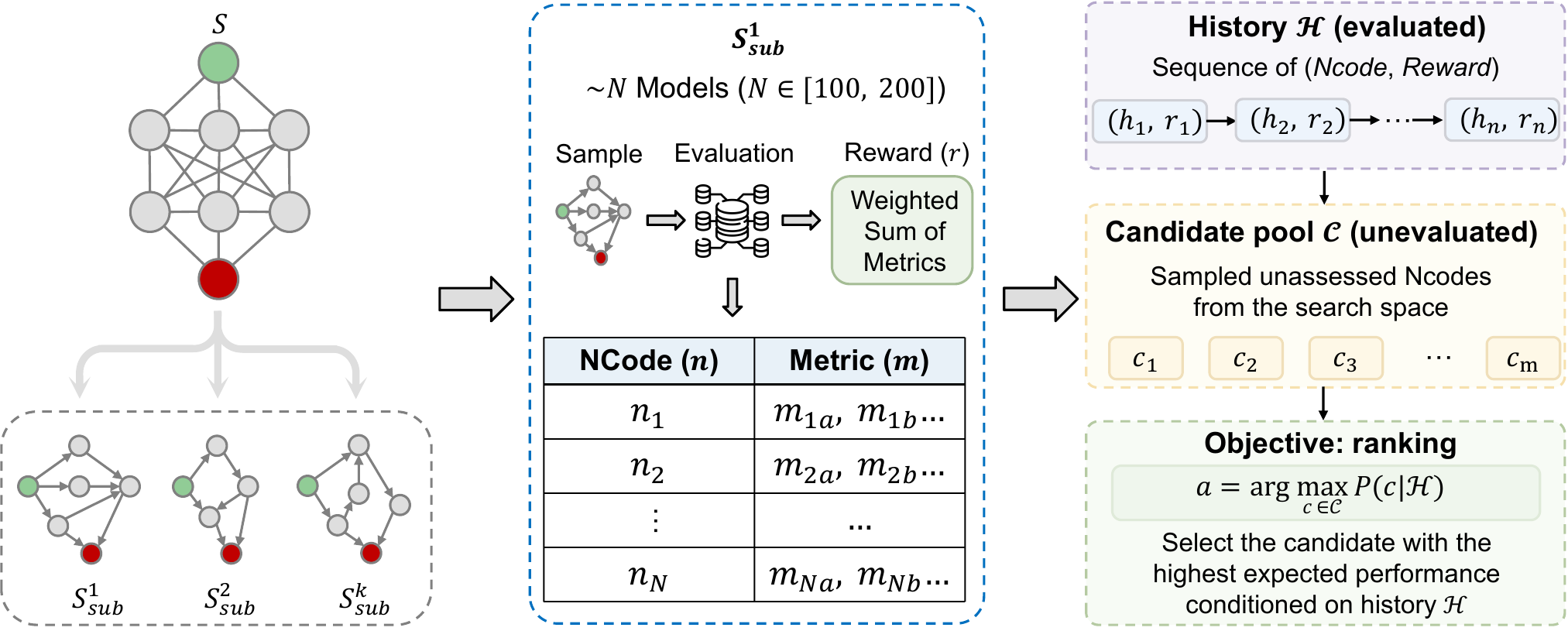} 
    \caption[]{Pipeline of the subspace-sampled data curation strategy. Given an entire architecture search space modeled as a directed acyclic graph (DAG), stochastic edge pruning is applied to generate structurally diverse, localized sub-spaces ($\mathcal{S}_{sub}$). Within each isolated sub-space, a subset of model architectures is sampled and evaluated to pair their unified NCode representations with corresponding performance metrics. To construct the training trajectories, these samples are partitioned into a sequential contextual history ($\mathcal{H}$) of evaluated trials and a pool of unassessed candidates ($\mathcal{C}$). This pipeline reframes the large-scale optimization problem into a tractable ranking task, training the model to select the candidate with the highest expected performance conditioned on historical context.}
    \label{fig:method_data_curation}
\end{figure*}

\subsection{Optimization Trajectory Curation}
\label{sec:trajectory_curation}

To effectively leverage LLMs as a general-purpose searcher for AutoML, we curated a dataset of optimization trajectories to capture the sequential, causal steps of iterative search processes. Specifically, each optimization trajectory comprises a sequence of AutoML configuration–reward pairs, with architectures represented by our unified NCode scheme. At each iteration, the search algorithm utilizes historically evaluated pairs as context to sample a new model architecture, which is subsequently evaluated to derive its performance metrics. A weighted sum of these metrics serves as the reward signal for the given architecture.

Instead of collecting AutoML joint optimization samples directly, which is time-consuming and computationally expensive, we sampled model-reward pairs from the established NAS dataset NAS-Bench-301 that includes approximately $10^{21}$ distinct model architectures with their performance on CIFAR-10 classification task. To augment the diversity of the architectural search space within our optimization data, we introduced a pruning-based architecture sampling strategy that generates varied sub-spaces from a single overarching search space. As illustrated in Fig.~\ref{fig:method_data_curation}, the comprehensive architecture search space $\mathcal{S}$ is modeled as a fully connected DAG. We independently prune each edge with a probability $p=0.5$. This stochastic independent pruning inherently defines a localized sub-search space $\mathcal{S}_{sub} \subset \mathcal{S}$. Iterating this independent pruning generates a multitude of distinct sub-search spaces, yielding a training dataset with highly varied structural distributions.

Given the immense scale of the architectural search space, which comprises approximately $10^{21}$ architectures, identifying a single optimal architecture to serve as an absolute ground truth is computationally intractable and poses a significant learning challenge for the model. Consequently, we reframed the objective of the LLM-based searcher from direct prediction to candidate ranking.  Within each isolated sub-space $\mathcal{S}_{sub}$, we randomly sample a set of $N$ models (where $N \in [100, 200]$) alongside their corresponding performance metrics. The contextual history $\mathcal{H}$, comprising a sequence of previously evaluated NCode representations and their rewards, is provided to the LLM-based searcher. The model is then tasked with selecting the most promising NCode from a candidate pool $\mathcal{C}$ of unassessed configurations sampled from the same sub-space.

Through this methodology, we curated a robust dataset comprising approximately 228,000 optimization trajectory instances. This extensive dataset boosts the LLM's capacity to optimize AutoML configurations in a domain-agnostic manner.

\subsection{Interpretability through LLM CoT Reasoning}
\label{sec:interpretability}
To unveil the LLM-based searcher's reasoning process for AutoML optimization, we utilized Group Relative Policy Optimization (GRPO)~\cite{shao2024deepseekmath} to fine-tune the Qwen3-8B~\cite{Yang2025Qwen3TR} LLM as self-explaining searcher which learns to navigate the optimal AutoML configuration guided by its own CoT reasoning. We then analyzed the CoT reasoning explained in natural language on the NAS-Bench-201~\cite{dong2020bench} and GenEval~\cite{Ghosh2023GenEvalAO} datasets, extracting its reasoning trajectories during the search phase for subsequent interpretability analysis.

\vspace{1em}
\noindent \textbf{Policy Optimization for AutoML Search.} We cast the LLM-based AutoML search as a Markov Decision Process (MDP) designed to optimize decision-making trajectories within the universal NCode space. For a given search context $x$ (comprising historical trials and current candidates), the searcher acts as a policy $\pi_\theta$ that generates a token sequence $y$ representing the combined reasoning (CoT) and the final configuration selection. To efficiently navigate this complex discrete search space without the memory overhead of an explicit value network, we implement Group Relative Policy Optimization (GRPO)~\cite{shao2024deepseekmath}. The reward for GRPO is a weighted sum of accuracy reward, verifying the selection of the optimal NCode candidate, and format reward, enforcing the structural integrity of the reasoning process required for interpretability analysis.

\begin{figure*}[!htbp]
    \centering
    \includegraphics[width=0.9\textwidth]{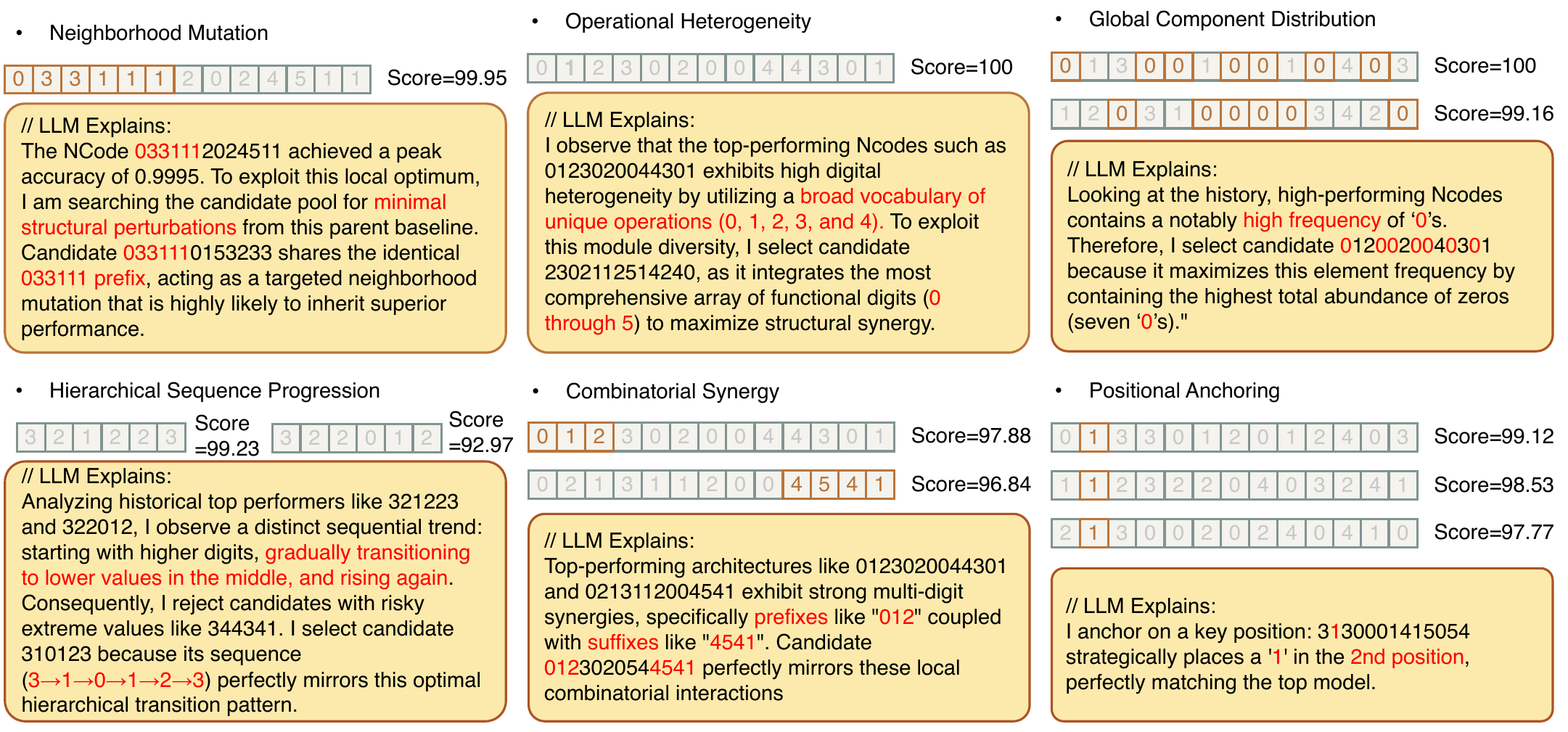} 
    \caption[]{Representative examples of the LLM’s reasoning process categorized by six distinct analytical patterns (e.g., Positional Anchoring, Combinatorial Synergy).}
    \label{fig:intepretability_examples}
\end{figure*}

\vspace{1em}
\noindent \textbf{Taxonomy of Chain-of-Thought Reasoning Behaviors} We employ LLM-based searcher with CoT capability to search on NAS-Bench-201 for image classification and GenEval for text-to-image generation. Experiments yielded 2,500 search steps complete with the model's intermediate reasoning process on each task. To systematically analyze the searcher's reasoning trajectories, we taxonomized the generated Chain-of-Thought outputs into six distinct behavioral patterns. The definitions of these patterns, as shown in Fig.~\ref{fig:intepretability_examples}, are as follows:

\begin{itemize}
    \item \textbf{Positional Anchoring}: This pattern occurs when the searcher explicitly identifies that the presence of a specific operational digit at a designated sequence position is strongly correlated with extreme (either high or low) performance outcomes. The LLM uses this single-point heuristic as an anchor for subsequent candidate selection.

    \item \textbf{Global Component Distribution}: This applies when the searcher bases its decision on the global distribution of operations. The model observes that the aggregate count of specific functional digits across the entire configuration sequence systematically correlates with overall model performance, regardless of their absolute positions.

    \item \textbf{Neighborhood Mutation}: We define this behavior as an evolutionary search strategy where the LLM selects a historically high-performing configuration as a baseline (``parent") and targets candidates that represent minimal structural perturbations (e.g., modifying only one or two sequence digits). This reflects an exploitation tactic focused on exploring the immediate local optimum neighborhood.

    \item \textbf{Combinatorial Synergy}: This describes instances where the reasoning process emphasizes the combinatorial interaction between multiple positions. The searcher identifies that specific multi-digit combinations (e.g., a coupled pair of operations at positions 3 and 5) yield a synergistic effect on performance that cannot be explained by individual digits alone.

    \item \textbf{Hierarchical Sequence Progression}: This categorizes behavior where the LLM detects monotonically increasing or decreasing numerical patterns within specific sub-regions of the configuration string. The model interprets these ordinal trends as reflections of smooth or abrupt hierarchical transitions in the network topology that systematically impact the target metric.

    \item \textbf{Operational Heterogeneity}: This pattern emerges when the searcher evaluates candidates based on their digital heterogeneity. By observing that high-performing models typically encompass a broader vocabulary of unique functional digits compared to homogeneous, underperforming ones, the LLM deliberately favors configurations that integrate a diverse array of numerical components.
\end{itemize}

\begin{algorithm}[ht]
\caption{LLM-based Zero-Shot AutoML Search Pipeline}
\label{alg:zero_shot_inference}
\textbf{Input:} Target task $\mathcal{T}$, search space $\mathcal{S}$, universal NCode encoder $\hat{f}$, proxy evaluator $E_{proxy}$, frozen LLM policy $\pi_{\theta}$, search budget $T$, buffer capacity $M$, initial samples $K_{init}$. \\
\textbf{Output:} The optimal AutoML configuration $a^*$. 
\begin{algorithmic}[1]
\STATE \textbf{Initialize} historical buffer $\mathcal{H} \leftarrow \emptyset$
\STATE \textbf{// Stage 1: Cold-Start Initialization}
\FOR{$i = 1$ to $K_{init}$}
    \STATE Sample configuration $a_i \sim \text{Random}(\mathcal{S})$
    \STATE Encode NCode sequence: $c_i \leftarrow \hat{f}(a_i)$
    \STATE Proxy evaluation on target task: $r_i \leftarrow E_{proxy}(a_i, \mathcal{T})$
    \STATE Update history: $\mathcal{H} \leftarrow \mathcal{H} \cup \{(c_i, r_i)\}$
\ENDFOR

\STATE \textbf{// Stage 2: Iterative Search Phase}
\FOR{$t = K_{init} + 1$ to $T$}
    \STATE Generate candidate pool $\mathcal{A}_{cand}$ via Evolutionary \& Random Search
    \STATE Encode pool: $\mathcal{C}_{cand} \leftarrow \{\hat{f}(a) \mid a \in \mathcal{A}_{cand}\}$
    
    \STATE \textbf{// Stage 3: LLM-Guided Selection}
    \STATE Construct prompt string $\mathcal{P}$ using $\mathcal{H}$ and $\mathcal{C}_{cand}$
    \STATE LLM reasoning and selection: $c^* \leftarrow \pi_{\theta}(\mathcal{P})$
    \STATE Decode to target configuration: $a^* \leftarrow \hat{f}^{-1}(c^*)$
    
    \STATE \textbf{// Stage 4: Proxy Evaluation}
    \STATE Execute early-stopping evaluation: $r^* \leftarrow E_{proxy}(a^*, \mathcal{T})$
    \STATE Update history: $\mathcal{H} \leftarrow \mathcal{H} \cup \{(c^*, r^*)\}$
    
    \STATE \textbf{// Stage 5: Bounded Buffer Management}
    \IF{$|\mathcal{H}| > M$}
        \STATE $\mathcal{H} \leftarrow \text{Top-}M \text{ elements in } \mathcal{H} \text{ ranked by } r$
    \ENDIF
\ENDFOR
\RETURN $a_{best} \leftarrow \arg\max_{(c, r) \in \mathcal{H}} r$ (Decoded back to config space)
\end{algorithmic}
\end{algorithm}

\begin{figure*}[!htbp]
    \centering
    \includegraphics[width=0.9\textwidth]{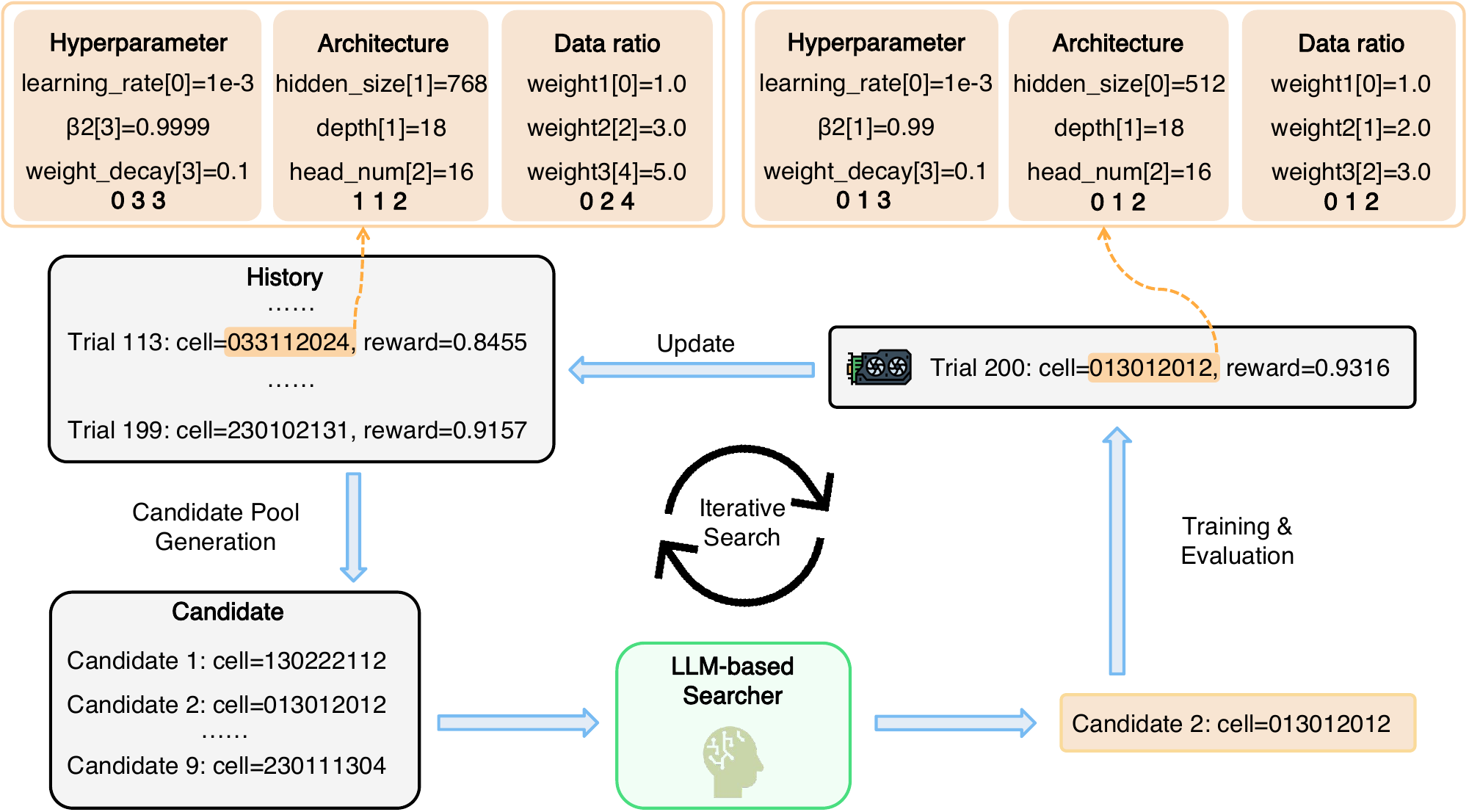} 
    \caption[]{Overview of our iterative LLM-based search framework. The training configurations encompassing hyperparameters, architectures, and data ratios are uniformly encoded as discrete numerical sequences named as NCodes. These NCodes are paired with their normalized performance metrics to serve as rewards, forming a continuously updated historical buffer of evaluated trials. During the search phase, a combination of an evolutionary algorithm and random search generates a pool of new candidate AutoML configurations. The LLM-based searcher analyzes the historical trials and identifies the most promising candidate from the candidate pool. The selected candidate undergoes training and evaluation, and the resulting performance metric is stored in the historical buffer. This iterative loop with NCodes is task- and domain-agnostic, enabling the LLM to zero-shot adapt to new AutoML domains without task-specific fine-tuning.}
    \label{fig:search_framework_overview}
\end{figure*}

\subsection{Zero-Shot Inference Pipeline}
\label{sec:inference}
Once the LLM-based searcher is trained on the curated trajectory dataset (as detailed in Sec.~\ref{sec:trajectory_curation}). It can be directly deployed to jointly optimize architectures, hyperparameters, and data ratios for unseen domains, novel search spaces, and distinct modalities without any task-specific gradient updates. This zero-shot inference pipeline acts as a closed-loop Markov Decision Process, illustrated in Fig.~\ref{fig:search_framework_overview}, and comprises the following core stages:

\begin{enumerate}
    \item \textbf{Cold-Start Initialization:}
    To initialize the historical context for the LLM, we employ a Random Search strategy to sample a small set of initial configurations (e.g., \(K_{\text{init}}=5\)). These configurations are mapped to their respective discrete sequences via the universal NCode $\hat{f}$. They are subsequently evaluated on the target task to obtain their performance metrics, which are aggregated and normalized into scalar rewards to form the initial historical buffer \(\mathcal{H}\).

    \item \textbf{Candidate Pool Generation:}
    At each search iteration, we generate a candidate pool \(\mathcal{C}\) of unassessed configurations. To ensure both exploration and exploitation, this pool is populated by applying an evolutionary algorithm (i.e., structural mutations on historically high-performing NCodes) alongside random sampling.

    \item \textbf{LLM-Guided Reasoning and Selection:}
    The current historical buffer \(\mathcal{H}\) (comprising evaluated NCode-reward pairs) and the candidate pool \(\mathcal{C}\) are integrated into a standardized prompt. The frozen LLM-based searcher processes this context, identifies and outputs the most promising candidate \(c^*\) from the pool.

    \item \textbf{Proxy Evaluation and Context Update:}
    To mitigate the prohibitive computational cost of fully training large-scale models during the search phase, we implement an early-stopping proxy evaluation protocol. The selected candidate is decoded back into the actual machine learning pipeline and trained for a minimal number of steps or epochs (e.g., 1 epoch for segmentation, 2500 steps for text-to-image synthesis). The derived surrogate reward \(r^*\) is then paired with its NCode \(c^*\) and appended to the historical buffer \(\mathcal{H}\).

    \item \textbf{Bounded Buffer Management:}
    To respect the LLM's maximum token context window and maintain high-quality context, the capacity of the historical buffer is bounded. Once the number of evaluated pairs exceeds a predefined threshold (e.g., \(M=200\)), a filtering mechanism is triggered to retain only the top-\(M\) performing configurations, discarding sub-optimal trials.
\end{enumerate}

This iterative cycle continues until a predefined computational budget is exhausted. The overall procedure is formalized in Algorithm~\ref{alg:zero_shot_inference}.

\section{Experiments}

\subsection{Overview}
To comprehensively evaluate the effectiveness, zero-shot transferability, scalability, and interpretability of our proposed LLM-based searcher, we conduct extensive empirical evaluations across four distinct machine learning scenarios. Specifically, we evaluate our framework on: (1) Cross-Domain Image Segmentation to test PEFT design and hyperparameter optimization under zero-shot domain shifts; (2) Cross-Modality Speaker Verification to assess cross-modal generalization from vision to audio; (3) Large-Scale Text-to-Image Synthesis to demonstrate joint architecture, hyperparameter, and data-ratio search at scale; and (4) NAS Benchmark (NAS-Bench-201) to verify standard neural architecture search performance. For each paradigm, we present the detailed experimental settings followed by the quantitative results. Furthermore, we investigate the interpretability of the LLM-based searcher's internal decision-making process, supported by thorough ablation studies and qualitative case studies. Below, we first outline the common implementation details before detailing each specific task.

\vspace{0.5em}
\noindent \textbf{Implementation Details}
\begin{itemize}
    \item \textbf{LLM-based Searcher Training:} To build the primary search agent, the LLaMA-3.1-8B base model~\cite{llama3modelcard} undergoes full-parameter supervised fine-tuning (SFT). The training is anchored on the curated dataset of approximately 228,000 optimization trajectory instances. To robustly capture expansive sequences of historical trial rewards alongside candidate configurations, we employ an extended context window with a strict token cutoff length of 20,480. The SFT process is executed for 3 epochs with a learning rate of $1.0 \times 10^{-5}$, regulated by a cosine learning rate scheduler following a 10\% warmup phase. To ensure memory efficiency during full-parameter optimization, training is distributed across a single node equipped with 8 GPUs using the DeepSpeed ZeRO-2 framework in bfloat16 mixed precision, configured with a per-device batch size of 1 and a gradient accumulation step of 2. 

    To analyze the searcher's internal decision-making process, we post-train a Qwen3-8B model using Group Relative Policy Optimization (GRPO). For this reinforcement learning stage, we subsample 2,000 representative trajectory instances from the primary 228,000-instance dataset and execute policy training for 3 epochs. To facilitate relative advantage estimation within the algorithmic framework of GRPO, the policy model is configured to generate a group of 8 independent candidate responses for each individual prompt sample. The computationally intensive training is scaled across a distributed 4-node cluster equipped with 8 GPUs per node (totaling 32 GPUs), running the DeepSpeed ZeRO-3 framework with CPU offloading in bfloat16 mixed precision. The execution configuration utilizes a per-device batch size of 1 matched with a gradient accumulation step of 4. To mitigate the immense token overhead of long-form Chain-of-Thought (CoT) generation while preserving search alignment, we employ Parameter-Efficient Fine-Tuning (PEFT) via Low-Rank Adaptation (LoRA) targeted at the attention projection modules (\texttt{q\_proj} and \texttt{v\_proj}). The maximum prompt and completion boundaries are constrained to 10,240 and 8,192 tokens respectively, allowing sufficient capacity for structural rationalization. A vLLM inference engine is colocated with the training framework to accelerate group-based relative reward sampling.

    \item \textbf{Search Phase:} To initialize the AutoML configuration search within our framework, we utilized Random Search to sample and evaluate five configurations. This generated an initial set of five NCode-reward pairs for the LLM-based searcher to analyze. In each subsequent iteration, the LLM proposed an NCode, which was then decoded into a specific training configuration. To ensure computational efficiency during searches for image segmentation and speaker verification, we applied early stopping after a single epoch and used the partially trained model to derive a surrogate performance measure. For large-scale text-to-image synthesis, we jointly searched model architectures, training hyperparameters, and data sampling ratios using a surrogate model scaled down to half the block depth and hidden dimension size of the target model, with training truncated at 2,500 steps. The overall reward was calculated as the weighted average of all metrics and subsequently normalized across all evaluated configurations. Notably, for metrics where lower values denote superior performance, their contribution to the final reward was negated. Throughout the search, we maintained a historical buffer of up to 200 NCode-reward pairs. Once the evaluated pairs exceeded this threshold, we filtered the context provided to the LLM, retaining only the top-200 performing configurations.

    \item \textbf{Retraining Phase:} Once the optimal configurations are identified by the LLM-based searcher during the early-stopping proxy evaluation, the selected configurations undergo a complete retraining phase for rigorous evaluation. Since these optimal configurations (encompassing architectures, hyperparameters, and data ratios) represent the direct outcomes of our search process, the specific retraining implementation details are inherently task-specific. Therefore, we detail the complete retraining setups within the respective Search Configuration sections of each task (Sections~\ref{sec:image_segment}--\ref{sec:text_to_image}).
\end{itemize}

\begin{table*}[!htbp]
\centering
\setlength{\tabcolsep}{4pt}
\renewcommand{\arraystretch}{1.1}
\caption{Joint optimization of model architecture and hyperparameters achieves state-of-the-art zero-shot transferability across diverse image segmentation domains. Results are reported as mean $\pm$ standard error over five independent runs. LM-Searcher performs architecture search in a zero-shot setting with respect to the target segmentation domain. Higher values indicate better performance for $S_{\alpha}$, $E_{\phi}$, $\mathrm{Jac}$, $F_{\beta}^{\omega}$, and IoU, whereas lower values indicate better performance for BER. Results for the architecture-only configuration, denoted as Ours (Arc)~\cite{hu2025lm}, are adopted from our preliminary conference version~\cite{hu2025lm} to serve as a baseline for demonstrating the benefits of joint optimization.}
\resizebox{0.95\textwidth}{!}{%
\begin{tabular}{lccccccccccc}
\hline
Method & \multicolumn{2}{c}{Kvasir-SEG} & \multicolumn{2}{c}{CVC-612} & ISIC2017 & \multicolumn{3}{c}{CAMO} & SBU & Leaf & Road \\
& $S_{\alpha}\uparrow$ & $E_{\phi}\uparrow$ & $S_{\alpha}\uparrow$ & $E_{\phi}\uparrow$ & $\mathrm{Jac}\uparrow$ & $S_{\alpha}\uparrow$ & $E_{\phi}\uparrow$ & $F_{\beta}^{\omega}\uparrow$ & $\mathrm{BER}\downarrow$ & $\mathrm{IoU}\uparrow$ & $\mathrm{IoU}\uparrow$ \\
\hline
BitFit~\cite{zaken2021bitfit}        & $90.8\pm0.57$ & $93.8\pm0.98$ & $89.0\pm0.40$ & $91.6\pm0.98$ & $76.4\pm0.45$ & $86.8\pm0.33$ & $90.7\pm0.28$ & $81.5\pm0.19$ & $3.2\pm0.13$ & $71.4\pm1.15$ & $60.6\pm0.15$ \\
Adapter~\cite{houlsby2019parameter}       & $91.2\pm0.23$ & $94.0\pm0.16$ & $89.3\pm0.43$ & $92.0\pm0.63$ & $76.7\pm0.66$ & $87.7\pm0.10$ & $91.3\pm0.40$ & $82.8\pm0.35$ & $2.8\pm0.09$ & $72.1\pm0.47$ & $61.5\pm0.11$ \\
VPT~\cite{jia2022visual}           & $91.5\pm0.23$ & $94.3\pm0.06$ & $91.0\pm0.94$ & $\underline{93.7\pm1.41}$ & $76.9\pm0.94$ & $87.4\pm0.60$ & $91.4\pm0.68$ & $82.1\pm0.75$ & $2.7\pm0.06$ & $73.6\pm0.26$ & $60.2\pm1.87$ \\
LST~\cite{sung2022lst}           & $89.7\pm0.25$ & $93.3\pm0.37$ & $89.4\pm0.37$ & $92.4\pm0.54$ & $76.4\pm1.05$ & $83.3\pm0.28$ & $88.0\pm0.23$ & $77.1\pm0.02$ & $3.2\pm0.01$ & $70.2\pm0.87$ & $60.2\pm0.26$ \\
SAM-Adapter~\cite{chen2023sam}   & $89.6\pm0.24$ & $92.5\pm0.10$ & $89.6\pm0.22$ & $92.4\pm1.06$ & $76.1\pm0.45$ & $85.6\pm0.26$ & $89.6\pm0.55$ & $79.8\pm0.89$ & $3.1\pm0.06$ & $71.4\pm0.20$ & $60.6\pm0.06$ \\
SSF~\cite{lian2022scaling}           & $91.3\pm0.87$ & $93.9\pm1.49$ & $89.6\pm0.37$ & $91.9\pm0.79$ & $76.6\pm0.19$ & $87.5\pm0.11$ & $91.4\pm0.16$ & $82.6\pm0.12$ & $3.2\pm0.05$ & $71.5\pm0.63$ & $61.6\pm0.03$ \\
LoRA~\cite{hu2021lora}          & $91.2\pm0.28$ & $93.8\pm0.22$ & $90.7\pm0.04$ & $92.5\pm0.41$ & $76.6\pm0.23$ & $88.0\pm0.24$ & $91.9\pm0.42$ & $82.8\pm0.16$ & $2.7\pm0.08$ & $73.7\pm0.20$ & $62.2\pm0.21$ \\
Conv-LoRA~\cite{zhong2024convolution}     & $91.9\pm0.64$ & $94.3\pm0.77$ & $90.1\pm0.31$ & $92.3\pm0.42$ & $\underline{77.0\pm0.51}$ & $88.6\pm0.12$ & $92.4\pm0.17$ & $83.1\pm0.19$ & $2.8\pm0.02$ & $73.9\pm0.20$ & $62.1\pm0.23$ \\
\textbf{Ours (Arc)~\cite{hu2025lm}}
               & $\underline{92.4\pm0.43}$ & $\underline{94.7\pm0.47}$ & $\underline{91.1\pm0.47}$ & $93.2\pm0.70$ & $\mathbf{77.6\pm0.40}$ & $\underline{88.9\pm0.04}$ & $\underline{92.7\pm0.15}$ & $\underline{83.9\pm0.32}$ & $\underline{2.6\pm0.04}$ &
               $\underline{74.2\pm0.28}$ & 
               $\underline{62.6\pm0.04}$ \\
\rowcolor{gray!10}
\textbf{Ours (Arc+Hyp)}
               & $\mathbf{92.7\pm0.43}$ & $\mathbf{95.5\pm0.55}$ & $\mathbf{92.6\pm0.56}$ & $\mathbf{94.3\pm0.80}$ & $\mathbf{77.6\pm0.57}$ & $\mathbf{89.3\pm0.26}$ & $\mathbf{93.0\pm0.12}$ & $\mathbf{83.9\pm0.32}$ & 
               $\mathbf{2.6\pm0.05}$ &
               $\mathbf{75.0\pm0.28}$ &
               $\mathbf{62.6\pm0.11}$ \\
\hline
\end{tabular}}
\label{tab:automl_segmentation}
\end{table*}

\subsection{Cross-Domain Image Segmentation}
\label{sec:image_segment}

\subsubsection{Experimental Setting}
\begin{itemize}
    \item \textbf{Datasets.} To evaluate the cross-task and cross-domain generalizability of our proposed LLM-based searcher, we utilized a diverse suite of image segmentation datasets encompassing multiple distinct domains. The search phase was conducted in the medical imaging domain using the Kvasir-SEG and CVC-612 datasets. Specifically, Kvasir-SEG and CVC-612 evaluate the model's capability in gastrointestinal polyp segmentation using endoscopic and colonoscopy imagery, respectively. To test zero-shot transferability, the optimal configuration discovered during search was evaluated on seven datasets across domains. In addition to the two datasets, these include ISIC2017, which focuses on the precise boundary delineation of skin lesions for melanoma detection; CAMO, which features natural imagery of camouflaged objects that artfully blend into their surroundings; SBU, which tasks the model with differentiating true structural boundaries from complex shadow masks in everyday scenes; Leaf, an agricultural dataset focused on delineating individual plant leaves from overlapping foliage backgrounds; and Road, which utilizes high-resolution remote sensing imagery to evaluate the extraction of continuous topological road networks across vast landscapes.
    
    \item \textbf{Search Configuration.} The search space comprises both model architectures and training hyperparameters. For architecture, we used the Segment Anything Model (SAM) as the base segmenter and searched the PEFT design by assigning the ranks of 32 LoRA modules in the ViT encoder from $\{3,6,12,24\}$, yielding a search space of approximately $10^{19}$ architectures. The hyperparameter search space incorporates the focal loss $\gamma$, weight decay, and learning rate. To efficiently navigate this expansive space, we implement an early-stopping proxy evaluation protocol during the search phase, where each candidate configuration is trained for only 1 epoch to derive a surrogate reward. The discovered optimal configuration is subsequently subjected to full retraining for 30 epochs, establishing a highly efficient search-to-full-training workload ratio of 1:30. Comprehensive details regarding the search space are provided in Appendix Table 1. For the subsequent retraining phase, we applied the identified best-performing AutoML configurations to each downstream dataset. Specifically, we fine-tuned the Segment Anything Model (SAM)~\cite{kirillov2023segment} using our optimal LoRA~\cite{hu2021lora} configuration. Training was conducted on a single GPU, utilizing a global batch size of 4, a learning rate of $1 \times 10^{-3}$, a weight decay of $3 \times 10^{-5}$, and a $\gamma$ value of 2.0.
    
    \item \textbf{Evaluation Metrics.} To comprehensively assess the zero-shot transferability of the searched configurations across diverse target domains, we employ a multi-faceted evaluation protocol tailored to the morphological characteristics of each dataset. For all metrics except the Balanced Error Rate (BER), higher values indicate superior performance.
    \begin{itemize}
        \item \textbf{Structural and Alignment Metrics:} For datasets requiring strict foreground object consistency and structural integrity (Kvasir-SEG, CVC-612, and CAMO), we utilize the S-measure ($S_\alpha$)~\cite{Fan2017StructureMeasureAN} and Enhanced-alignment measure ($E_\phi$)~\cite{Fan2018EnhancedalignmentMF}.
        \item \textbf{Spatial Overlap and Region Metrics:} For tasks where exact spatial overlap is the most critical success factor (ISIC2017, Leaf, and Road), we rely on the widely adopted Jaccard Index (Jac). Additionally, the weighted F-measure ($F_\beta^\omega$) is evaluated across multiple datasets to adaptively emphasize errors in highly salient regions.
        \item \textbf{Balanced Error Metrics:} The Balanced Error Rate (BER) is specifically applied to evaluate the performance on the SBU dataset. It ensures an unbiased assessment for shadow detection by equally weighting the classification error rates of both shadow and non-shadow regions.
    \end{itemize}
    Detailed mathematical formulations for all the aforementioned evaluation metrics are provided in Appendix Section 2.
\end{itemize}

\subsubsection{Experimental Results}
As detailed in Table~\ref{tab:automl_segmentation}, the optimal configuration discovered by our LLM-based searcher, denoted as Ours (Arc+Hyp), establishes a new state-of-the-art across all evaluated datasets and metrics. Notably, despite the search phase being exclusively anchored in the medical imaging domain, the resulting configuration exhibits competitive zero-shot generalizability, successfully transferring to natural (CAMO, SBU), agricultural (Leaf), and remote sensing (Road) imagery. It consistently outperforms both domain-agnostic PEFT baselines (e.g., LoRA~\cite{hu2021lora}, Adapter, VPT~\cite{jia2022visual}) and task-specific adaptations (e.g., SAM-Adapter~\cite{chen2023sam}, Conv-LoRA~\cite{zhong2024convolution}). For instance, on the CVC-612 dataset, our method achieves an $S_{\alpha}$ of 92.6 and an $E_{\phi}$ of 94.3, yielding absolute improvements of 2.5 and 2.0 percentage points, respectively, over the strongest baseline, Conv-LoRA~\cite{zhong2024convolution}. Furthermore, ablation studies validate our joint search paradigm: simultaneously searching architecture and hyperparameters (Ours (Arc+Hyp)) consistently surpasses architecture-only search (Ours (Arc)) across all benchmarks.

\begin{table*}[t]
\centering
\caption{\protect\internallinenumbers{The proposed searcher demonstrates superior zero-shot cross-modality generalization on the 3D-Speaker verification benchmark. Lower values indicate better performance for all metrics. ``Average'' denotes the mean EER across the cross-device, cross-distance, and cross-dialect tracks.}}
\begin{tabular}{lcccccccc}
\hline
Method & \multicolumn{2}{c}{Multi-Device} & \multicolumn{2}{c}{Multi-Distance} & \multicolumn{2}{c}{Multi-Dialect} & \multicolumn{2}{c}{Average} \\
 & EER(\%)$\downarrow$ & MinDCF$\downarrow$ & EER(\%)$\downarrow$ & MinDCF$\downarrow$ & EER(\%)$\downarrow$ & MinDCF$\downarrow$ & EER(\%)$\downarrow$ & MinDCF$\downarrow$ \\
\hline
Res2Net~\cite{gao2019res2net}      & 8.03 & 0.707 & 9.67 & 0.781 & 14.11 & 0.920 & 10.60 & 0.802 \\
ResNet34~\cite{koonce2021resnet}     & 7.29 & 0.689 & \textbf{8.98} & 0.762 & 12.81 & 0.906 & 9.69 & 0.786 \\
ECAPA-TDNN~\cite{desplanques2020ecapa}   & 8.55 & 0.728 & 12.15 & 0.814 & 12.24 & 0.915 & 10.98 & 0.819 \\
ERes2Net~\cite{chen2023enhanced}     & 7.12 & 0.657 & 9.82 & 0.749 & 12.10 & 0.866 & 9.68 & 0.757 \\
CAM++~\cite{wang2023cam++}        & 7.17 & 0.669 & 9.84 & 0.722 & \underline{11.78} & 0.844 & 9.60 & 0.745 \\
ERes2NetV2~\cite{chen20253d}   & \textbf{6.82} & \textbf{0.628} & 9.35 & 0.723 & 11.93 & \underline{0.826} & \underline{9.37} & \underline{0.726} \\
OptFormer~\cite{chen2022towards}    & 7.05 & 0.668 & 11.54 & \underline{0.802} & 12.01 & 0.842 & 10.20 & 0.771 \\
\rowcolor{gray!10}
Ours & \underline{6.88} & \underline{0.629} & \underline{9.10} & \textbf{0.697} & \textbf{11.66} & \textbf{0.819} & \textbf{9.21} & \textbf{0.715} \\
\hline
\end{tabular}%
\label{tab:automl_sv_results}
\end{table*}

\subsection{Cross-Modality Speaker Verification}

\subsubsection{Experimental Setting}
\begin{itemize}
    \item \textbf{Datasets.} We validated the cross-modality generalization capability of our LLM-based searcher on the audio modality evaluation suite 3D-Speaker-Toolkit~\cite{chen20253d}. The 3D-Speaker corpus is a large-scale dataset containing over 10,000 speakers, which are partitioned into training and test sets. We evaluated the models across three challenging tracks: multi-device, multi-distance, and multi-dialect. The evaluation split comprises 240 unseen speakers, 18,782 utterances, 11 dialects, and 15.42 hours of speech.
    
    \item \textbf{Search Configuration.} Our framework navigates a comprehensive search space that comprises both model architectures and training hyperparameters for the speaker verification task. The architectural search space dictates the model's capacity and feature resolution, specifically targeting the Mel-filterbank input dimensions, the base channel width of the ERes2NetV2 backbone, and the final speaker embedding size. Concurrently, the hyperparameter search space includes the data augmentation probability, learning rate, optimizer momentum, and weight decay. To maintain computational efficiency across this joint space, we implement an early-stopping proxy evaluation protocol during the search phase, truncating each candidate's training at only 1 epoch to derive a surrogate reward. The discovered optimal configuration is subsequently subjected to full retraining for a total of 77 epochs—comprising 70 epochs of standard training followed by a 7-epoch large-margin fine-tuning stage—establishing a highly efficient search-to-full-training workload ratio of approximately 1:77. The exact search space for each design choice is detailed in Appendix Table 1. For the retraining phase, the optimal model architecture was trained on the 3D-Speaker-Toolkit dataset for 70 epochs. We employed a cosine learning rate scheduler with an initial learning rate of 0.3, a weight decay of $5 \times 10^{-5}$, a momentum of 0.85, and a data augmentation probability of 0.4. The resulting model was subsequently fine-tuned utilizing a large-margin strategy for an additional seven epochs, maintaining the same hyperparameter settings.
    
    \item \textbf{Evaluation Metrics.} We assessed the performance of the discovered configurations utilizing two standard metrics within the speaker verification domain: the Equal Error Rate (EER) and the minimum of the normalized detection cost function (minDCF). For both metrics, lower values signify superior detection and recognition performance. The optimal configuration was fully retrained and evaluated across five independent runs, from which we report the mean and standard deviation.
\end{itemize}

\subsubsection{Experimental Results}
As reported in Table~\ref{tab:automl_sv_results}, our approach achieves an Equal Error Rate (EER) of 9.21 and a Minimum Detection Cost Function (minDCF) of 0.715 on average across all tracks, notably surpassing the performance of the current state-of-the-art manually-tuned ERes2NetV2. Furthermore, our framework significantly outperforms the competitive AutoML baseline, OptFormer~\cite{chen2022towards}\footnote{Since the original OptFormer is not natively designed for architecture search tasks, we adapted it for a fair comparison by training a model with an identical parameter count, architecture, and training data, strictly following the data processing protocols detailed in the original OptFormer paper.}, which attains a higher average EER of 10.20 and a minDCF of 0.771. This demonstrates that our method can effectively generalize across modalities without any further fine-tuning, considering the LLM-based searcher was trained solely on image modality data and zero-shot transferred to the 3D-Speaker audio dataset~\cite{chen20253d}.

\begin{table*}[t]
\centering
\setlength{\tabcolsep}{4.5pt} 
\caption{\protect\internallinenumbers The optimized Diffusion Transformer achieves state-of-the-art compositional text-to-image alignment on the GenEval benchmark with high parameter efficiency. Arrows ($\uparrow$) indicate that higher scores correspond to better performance.}
\begin{tabular}{@{} l l c c c c c c c @{}}
\toprule
Model & \# Params & Single object $\uparrow$ & Two object $\uparrow$ & Counting $\uparrow$ & Colors $\uparrow$ & Position $\uparrow$ & Attribute binding $\uparrow$ & Overall $\uparrow$ \\
\midrule
SDv2.1~\cite{rombach2022high}       & 0.9B  & 0.98 & 0.51 & 0.44 & 0.85 & 0.07 & 0.17 & 0.50 \\
SDXL~\cite{podell2024sdxl}         & 2.6B  & 0.96 & 0.74 & 0.39 & 0.85 & 0.15 & 0.23 & 0.55 \\
SD3-medium~\cite{esser2024scaling}   & 2B    & 0.99 & 0.94 & 0.72 & 0.89 & 0.33 & 0.60 & 0.74 \\
FLUX.1-dev~\cite{flux2024}   & 12B   & 0.99 & 0.81 & 0.79 & 0.74 & 0.20 & 0.47 & 0.67 \\
Emu3-Gen~\cite{wang2024emu3} & 8B & 0.98 & 0.71 & 0.34 & 0.81 & 0.17 & 0.21 & 0.54 \\ 
JanusFlow~\cite{ma2025janusflow}        & 2B    & 0.97 & 0.59 & 0.45 & 0.83 & 0.53 & 0.42 & 0.63 \\
Janus-Pro-7B~\cite{chen2025janus}     & 7B    & 0.99 & 0.89 & 0.59 & 0.90 & 0.79 & 0.66 & 0.80 \\
OmniGen~\cite{xiao2025omnigen}      & 3.8B  & 0.98 & 0.84 & 0.66 & 0.74 & 0.40 & 0.43 & 0.68 \\
BAGEL~\cite{deng2025emerging}        & 7B    & 0.98 & 0.95 & 0.84 & 0.95 & 0.78 & 0.77 & 0.88 \\
Lumina-Image 2.0~\cite{qin2025lumina} & 2.6B    & -    & 0.87 & 0.67 & -    & -    & 0.62 & 0.73 \\
HiDream-I1-Full~\cite{cai2025hidream}  & 17B    & \textbf{1.00} & \textbf{0.98} & 0.79 & 0.91 & 0.60 & 0.72 & 0.83 \\
Qwen-Image~\cite{wu2025qwen}   & 20B   & 0.99 & 0.92 & 0.89 & 0.88 & 0.76 & 0.77 & 0.87 \\
GPT Image 1 [High]~\cite{openai2025gpt-image-1}  & -    & 0.99 & 0.92 & 0.85 & 0.92 & 0.75 & 0.61 & 0.84 \\
Seedream 3.0~\cite{gao2025seedream}     & -    & 0.99 & 0.96 & \textbf{0.91} & 0.93 & 0.47 & 0.80 & 0.84 \\
OptFormer~\cite{chen2022towards}    & 4B    & 0.99 & 0.90 & 0.76 & 0.92 & 0.73 & 0.74 & 0.84 \\
\rowcolor[gray]{.9}
Ours  & 1.1B  & 0.99 & 0.96 & 0.85 & \textbf{0.96} & \textbf{0.86} & \textbf{0.85} & \textbf{0.91} \\
\bottomrule
\end{tabular}%
\label{tab:geneval_results}
\end{table*}

\begin{figure}[!htbp]
    \centering
    \includegraphics[]{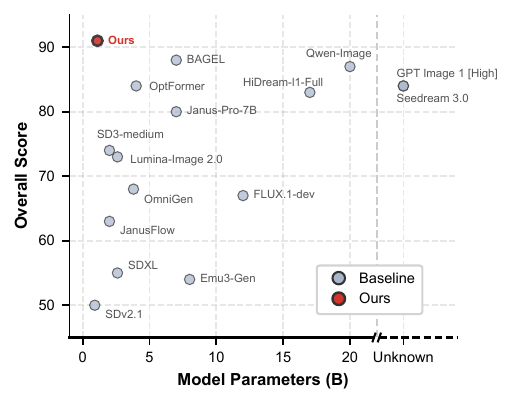} 
    \caption[]{{Performance comparison of our image generative model using configurations optimized by the proposed LLM-based searcher against baseline methods on the GenEval benchmark.} Our method achieves the highest overall score while using significantly fewer parameters than competing state-of-the-art models.}
    \label{fig:geneval_model_comparison}
\end{figure}

\subsection{Large-Scale Text-to-Image Synthesis}
\label{sec:text_to_image}
\subsubsection{Experimental Setting}
\begin{itemize}
    \item \textbf{Heterogeneous Training Corpus.} To train the DiT model, we constructed a diverse image-text corpus and dynamically search for the optimal data distribution ratios during training. This corpus amalgamates several high-quality datasets, each providing image-caption pairs with unique visual-semantic priors:
    \begin{itemize}
        \item \textit{BLIP3o} (Short, Long, JourneyDB, and 60k variants)~\cite{Chen2025BLIP3oAF}: Provides highly descriptive, synthetically enriched captions that enhance fine-grained semantic alignment.
        \item \textit{LAION-high-resolution}~\cite{Schuhmann2022LAION5BAO}: A large-scale, web-scraped dataset ensuring broad coverage of real-world concepts and natural image distributions.
        \item \textit{Midjourney-23m-2}: Comprises high-fidelity, artistically curated synthetic images to elevate the overall aesthetic quality of the generated outputs.
        \item \textit{X2I2}~\cite{Wu2025OmniGen2ET}: A dataset designed to improve complex instruction-guided generation.
        \item \textit{Echo-4o}~\cite{Ye2025Echo4oHT}: A synthetic dataset generated by GPT-4o, providing clean supervision to complement rare scenarios and complex compositions less represented in real-world data.
    \end{itemize}

    \item \textbf{Search Configuration.} Our LLM-based searcher navigates a unified configuration space to jointly optimize the model architecture, training hyperparameters, and data sampling ratios. Architecturally, we adopt the MM-DiT as the foundational backbone, searching over the hidden dimension size, as well as the depths of the single- and double-stream blocks. The hyperparameter search space encompasses the learning rate, weight decay, and the optimizer parameters $\beta$ and $\epsilon$. Finally, to optimize the training data composition, the sampling weight for each dataset during training is independently tuned within a range of 1.0 to 6.0. To scale this framework to the immense computational demands of large-scale pre-training, we execute the search phase using a surrogate model scaled down to half the block depth and hidden dimension size of the target model, with training truncated at 2,500 steps. The optimal configuration is then scaled up to the full 1.3-billion-parameter architecture and trained for a total of 100,000 steps across three distinct resource-intensive stages, establishing a highly efficient search-to-full-training step ratio of approximately 1:40. Detailed descriptions of the search space are provided in Appendix Table 3. For the full retraining phase, we conducted a large-scale pre-training experiment, training a 1.3-billion-parameter model across 128 GPUs for approximately seven days. The training process consisted of three stages. In the first stage, all training images were resized to a $256 \times 256$ resolution. The model was trained for 85,000 steps on BLIP3o-JourneyDB, BLIP3o-Long, BLIP3o-Short, LAION-high-resolution, Midjourney-23m-2, and X2I2, with data sampling ratios of 1.0, 2.0, 3.0, 5.0, 1.0, and 4.0, respectively. We utilized a learning rate of $1 \times 10^{-3}$, a $\beta_2$ parameter of 0.99, an $\epsilon$ of $1 \times 10^{-15}$, and a weight decay of 0.1. In the second stage, we scaled the image resolution to $512 \times 512$ and continually trained the model for 5,000 steps using identical hyperparameters and data ratios. Finally, the model was fine-tuned on the BLIP3o-60k and Echo-4o datasets for 10,000 steps.
    
    \item \textbf{Evaluation Metrics.} We benchmarked the optimized DiT model using GenEval~\cite{Ghosh2023GenEvalAO} to rigorously assess fine-grained compositional alignment at the instance level. We systematically evaluated the model across fundamental visual primitives, including single/multi-object rendering, object counting, spatial positioning, coloring, and attribute binding. The overall GenEval score is reported as the unweighted macro-average of the accuracies across these distinct sub-tasks, ensuring equal evaluation weight for each compositional skill. For evaluation, we generate four samples per prompt using an inference step number of 50 and a Classifier-Free Diffusion Guidance (CFG) scale of 5.0.
\end{itemize}

\subsubsection{Experimental Results}
Table~\ref{tab:geneval_results} illustrates the performance comparison between our optimized DiT model and established text-to-image synthesis works on the GenEval~\cite{Ghosh2023GenEvalAO} metric. Notably, our approach achieves the highest GenEval score of 0.91 compared with existing state-of-the-art methods. We also benchmarked our approach against another AutoML method, OptFormer~\cite{chen2022towards}, within the identical search space, outperforming it by an absolute margin of 0.07. Moreover, our model demonstrates remarkable parameter and data efficiency: despite comprising only 1.3 billion parameters and being trained on merely 50 billion tokens, it exceeds the performance of the 20-billion-parameter Qwen-Image model by 0.04. Furthermore, it surpasses the BAGEL model, which relies on 7 billion parameters and a massive 5 trillion training tokens, by a margin of 0.03 (Fig.~\ref{fig:geneval_model_comparison}). These findings underscore the robust generalization capability and scalability of our method in large-scale machine learning training scenarios.

\subsection{NAS Benchmark}

\subsubsection{Experimental Setting}
\begin{itemize}
    \item \textbf{Datasets.} While traditional NAS methods are highly specialized for image classification tasks, we benchmark our approach on NAS-Bench-201 to verify that our domain-agnostic LLM-based searcher matches their performance in standard image classification, serving as a prerequisite for our open-world AutoML evaluations. Every architecture within the NAS-Bench-201 search space has been fully trained and evaluated across three distinct image classification datasets: CIFAR-10, CIFAR-100, and ImageNet-16-120. By querying this predefined dataset, we bypass the immense computational overhead of repetitively training candidate architectures from scratch.
    
    \item \textbf{Search Space Configuration.} NAS-Bench-201 offers an algorithm-agnostic, fixed search space based on a neural cell topology represented as a DAG. The cell architecture comprises four nodes and five predefined operation options (zeroize, skip connection, 1-by-1 convolution, 3-by-3 convolution, and 3-by-3 average pooling), which yields a comprehensive search space of 15,625 unique neural architecture candidates.
    
    \item \textbf{Evaluation Protocol.} We executed five independent search trials of 200 steps per dataset. The reported metrics represent the average classification accuracy across the five top-tier architectures, each selected based on the highest validation accuracy attained during its individual run.
\end{itemize}

\subsubsection{Experimental Results}
As illustrated in Table~\ref{tab:nas_bench_201}, our proposed LLM-based searcher demonstrates highly competitive performance across the CIFAR-10, CIFAR-100, and ImageNet-16-120 datasets without any task-specific fine-tuning. Although our framework is designed as a general-purpose, open-world optimizer rather than an architecture-specialized NAS algorithm, it yields performance highly comparable to state-of-the-art specialized methods. Specifically, on ImageNet16-120, our zero-shot searcher achieves a test accuracy of 46.51\%, noticeably outperforming the top conventional NAS baseline DiNAS~\cite{asthana2024multi} (45.41\%) as well as earlier LLM-based searchers such as LLMatic~\cite{nasir2024llmatic} (45.87\%) and GENIUS~\cite{zheng2023can} (44.96\%), while performing closely on par with the specialized CoLLM-NAS~\cite{li2026collm} (46.79\%). These results confirm that our task-agnostic paradigm retains strong search efficiency on standard architectural benchmarks, providing a solid prerequisite for multi-modal open-world AutoML tasks.

\begin{table}[t]
    \centering
    \caption{Performance comparison on the NAS-Bench-201 benchmark. For each of the five experimental runs, we select the architecture with the highest validation accuracy and report its corresponding average test accuracy. The evaluation results of ours are adapted from our preliminary conference paper~\cite{hu2025lm}.}
    \resizebox{1.0\linewidth}{!}{
        \begin{tabular}{l|c|c|c|c|c|c}
          \toprule
          \multirow{2}{*}{Method} & \multicolumn{2}{c|}{CIFAR10} & \multicolumn{2}{c|}{CIFAR100} & \multicolumn{2}{c}{ ImageNet16-120} \\

          & Validation & Test & Validation & Test & Validation & Test  \\
          \midrule
          
          \rowcolor{gray!10}
          \multicolumn{7}{l}{\textit{Conventional}} \\
          
          Random Search~\cite{li2020random} & \underline{91.12} & \underline{93.89} & \underline{72.08} & \underline{72.07} & \underline{45.87} & \textbf{45.98} \\
          DARTS \citep{liu2018darts} & 39.77 & 54.30 & 38.57 & 15.61 & 18.87 & 16.32 \\
          PC-DARTS~\cite{xu2019pc} & 89.96 & 93.41 & 67.12 & 67.48 & 40.83 & 41.31 \\
          SNAS~\cite{xiesnas} & 90.10 & 92.77 & 69.69 & 69.34 & 42.84 & 43.16 \\
          SGNAS \citep{huang2021searching} & 90.18 & 93.53 & 70.28 & 70.31 & 44.65 & 44.98 \\
          DiNAS \citep{asthana2024multi} & \textbf{91.61} & \textbf{94.37} & \textbf{73.49} & \textbf{73.51} & \textbf{46.66} & \underline{45.41} \\
          \midrule
          
          \rowcolor{gray!10}
          \multicolumn{7}{l}{\textit{LLM-based}} \\
          GENIUS~\citep{zheng2023can} & 91.07 & 93.79 & 70.96 & 70.91 &  45.29 & 44.96  \\
          LLMatic~\cite{nasir2024llmatic} & 91.42 & \underline{94.26} & 71.41 & 71.62 & 44.98 & 45.87 \\
          CoLLM-NAS~\cite{li2026collm} & \textbf{91.59} & \textbf{94.37} & \textbf{73.44} & \textbf{73.44} & \textbf{46.62} & \textbf{46.79} \\
          \rowcolor[gray]{.9}
          Ours & \underline{91.52} & 94.20 & \underline{72.82} & \underline{72.96} & \underline{46.48} & \underline{46.51}  \\
          \bottomrule
        \end{tabular}
    }
    \label{tab:nas_bench_201}
\end{table}

\begin{figure*}[!htbp]
    \centering
    \includegraphics[width=0.95\textwidth]{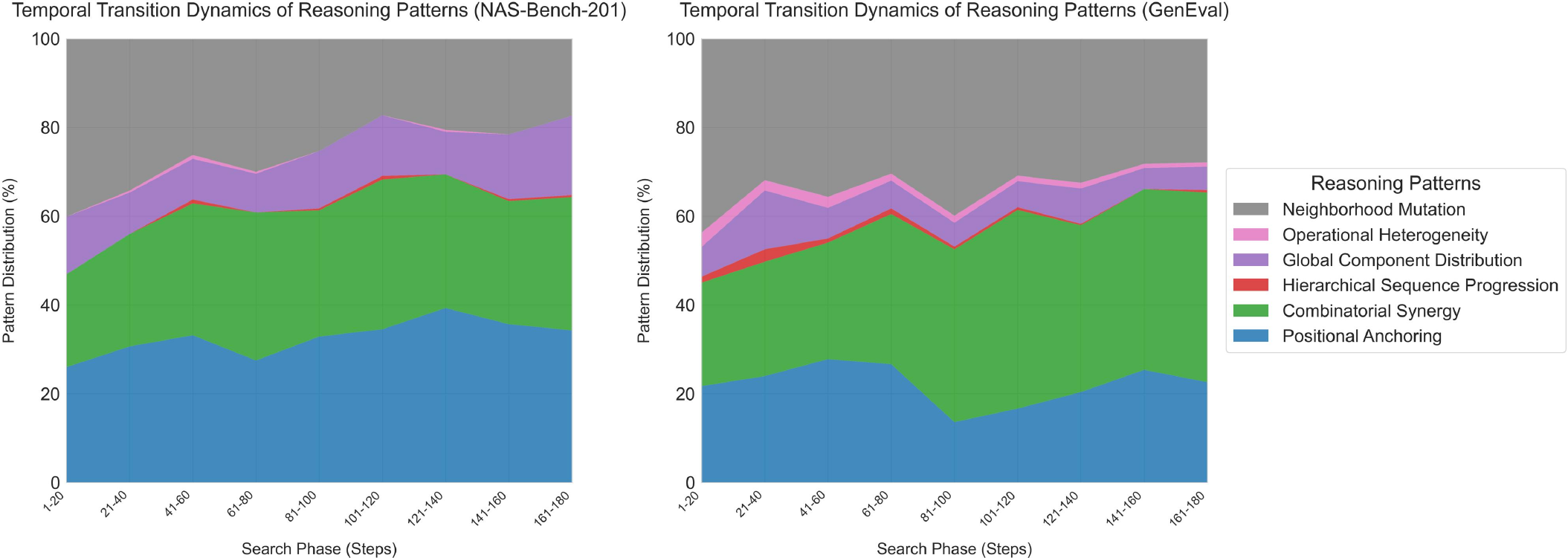} 
    \caption[]{Temporal transition dynamics of the reasoning pattern distributions throughout the search phase (steps 0–180). The stacked area charts illustrate how the LLM dynamically shifts its reliance on different reasoning strategies over time across the NAS-Bench-201 (left) and GenEval (right) benchmarks.}
    \label{fig:stacked_area_chart}
\end{figure*}

\subsection{Interpretability of LLM-based Searcher’s Reasoning Rationales}\label{result:interpretability}

\subsubsection{Experimental Setting}
To unveil the proposed LLM-based searcher's reasoning process in optimizing heterogeneous AutoML configurations, we post-train a Qwen3-8B model with CoT capabilities using our curated optimization trajectory dataset with GRPO (Fig.~\ref{fig:interpretability_pipeline}). Crucially, as validated in our ablation studies (Section~\ref{sec:ablations}), this Qwen3-8B-based searcher exhibits performance parity with our primary LLaMA-3.1-8B backbone, establishing it as a faithful proxy to decode the framework's decision-making logic. We employed the LLM-based searcher with CoT capability to search for image classification on NAS-Bench-201 and text-to-image synthesis on GenEval, generating optimization trajectories alongside natural language reasoning processes for explainability. These ``reasoning processes'' were then taxonomized using Gemini-2.5-Pro to identify six major optimization strategies learned by the LLM-based searcher (Fig.~\ref{fig:stacked_area_chart}): Positional Anchoring (e.g., fixing high-performing digit at critical position), Combinatorial Synergy (e.g., combinational effect of digits across positions), Neighborhood Mutation (e.g., making slight perturbations on evaluated NCode), Global Component Distribution (e.g., tracking the occurrence of specific digit), Hierarchical Sequence Progression (increasing/decreasing trends at local regions) and Operational Heterogeneity (diversity of digits in NCodes).

\subsubsection{Experimental Results}
Subsequent analysis of strategy distributions, state transitions, and temporal dynamics reveals that the LLM intrinsically develops a sophisticated exploration-exploitation paradigm, progressively shifting from broad structural exploration to the greedy exploitation of key architectural anchors.

Analyzing the temporal evolution of these reasoning patterns unveils a reasonable search logic rather than a random application of strategies (Fig.~\ref{fig:stacked_area_chart}). Initially, the searcher predominantly employs Neighborhood Mutation to navigate the configuration space, selecting candidate NCodes that introduce localized structural perturbations to historically high-performing baselines. As the search progresses, the probability of utilizing this Neighborhood Mutation gradually diminishes, giving way to exploitative patterns that leverage historical context. Crucially, this transition adapts to the complexity of the target task. On the simpler NAS-Bench-201 image classification benchmark, the searcher gradually shifts to the Positional Anchoring strategy, indicating a learned tendency to fix specific, high-yielding modules at critical topological positions. Conversely, on the highly complex GenEval text-to-image synthesis task, the model gradually shifts toward Combinatorial Synergy. By focusing on multi-digit patterns within NCodes, it demonstrates an ability to untangle the intricate combinatorial effects of heterogeneous configurations. 

Furthermore, the trajectory of Global Component Distribution, wherein the searcher evaluates configurations based on the macro-level frequency of specific operational digits rather than their absolute positions, reveals distinct adaptations. In the homogeneous NAS-Bench-201 search, the utilization of this strategy progressively intensifies, scaling up from an initial 10\% to nearly 20\%. In sharp contrast, for the heterogeneous GenEval task, this pattern is kept consistently subdued, hovering steadily at a probability of approximately 5\%. This divergence offers compelling evidence of the system's open-world generalizability: while the model learns to capitalize on global frequency heuristics as a stable descriptor for homogeneous model architecture search, it correctly recognizes that joint optimization across architecture, data, and hyperparameters demands a strict departure from simple global counting. Ultimately, these dynamics demonstrate that the LLM-based searcher does not merely memorize patterns; rather, it effectively navigates search spaces by adaptively combining different strategies to pinpoint optimal configurations.

\begin{figure*}[!htbp]
    \centering
    \includegraphics[width=0.95\textwidth]{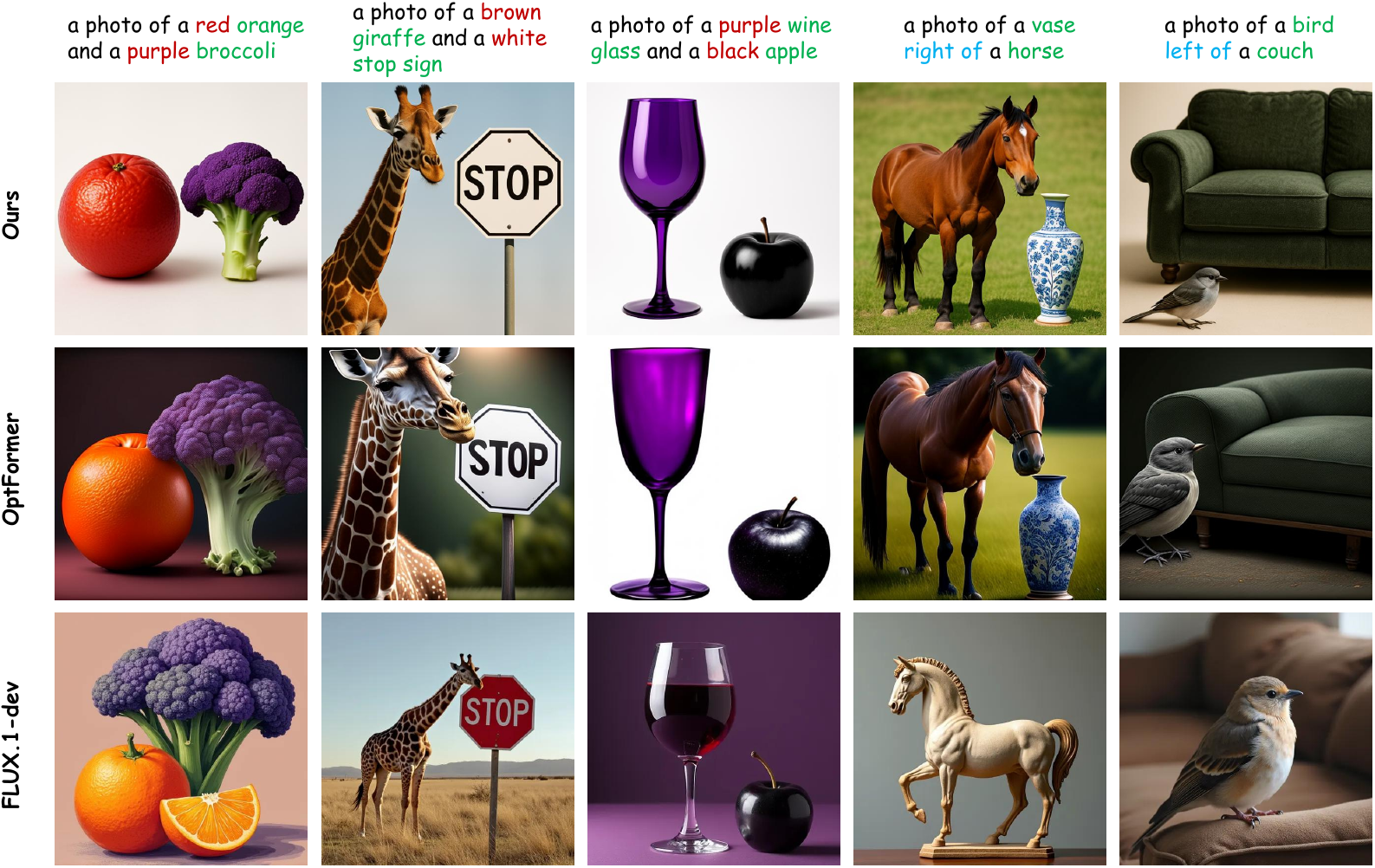} 
    \caption[]{Qualitative comparison of text-to-image synthesis results on the GenEval benchmark. We compare our optimized model against FLUX.1-dev and OptFormer across complex prompts requiring precise coloring and attribute binding (Columns 1–3), as well as spatial reasoning (Columns 4–5).}
    \label{fig:geneval_case_study}
\end{figure*}

\vspace{2em}

\begin{table*}[htbp]
\centering
\caption{\protect\internallinenumbers{Ablation performance comparison of different methods on the GenEval benchmark.}}
\resizebox{\textwidth}{!}{%
\begin{tabular}{@{} l l c c c c c c c @{}}
\toprule
Method & \# Params & Single object $\uparrow$ & Two object $\uparrow$ & Counting $\uparrow$ & Colors $\uparrow$ & Position $\uparrow$ & Attribute binding $\uparrow$ & Overall $\uparrow$ \\
\midrule
Only search architecture  & 4.1B & 0.98 & 0.93 & 0.78 & 0.94 & 0.76 & 0.77 & 0.86 \\
Search hyperparameter and data ratio & 1.8B & \textbf{0.99} & 0.95 & 0.79 & 0.92 & 0.83 & 0.86 & 0.89 \\
w/o searched hyperparameter and data ratio & 1.1B & \textbf{0.99} & 0.89 & 0.68 & 0.91 & 0.76 & 0.72 & 0.82 \\
w/o searched architecture & 1.8B & \textbf{0.99} & 0.91 & 0.75 & 0.92 & 0.76 & 0.73 & 0.84 \\
\rowcolor[gray]{.9}
Ours  & 1.1B  & \textbf{0.99} & \textbf{0.96} & \textbf{0.85} & \textbf{0.96} & \textbf{0.86} & 0.85 & \textbf{0.91} \\
\bottomrule
\end{tabular}%
}
\label{tab:geneval_ablation}
\end{table*}

\vspace{2em}

\begin{table*}[htbp]
    \centering
    \caption{\protect\internallinenumbers{Ablation performance on NAS-Bench-201 using different models and training strategies.}}
    \resizebox{0.9\textwidth}{!}{%
        \begin{tabular}{l|c|c|c|c|c|c}
          \toprule
          \multirow{2}{*}{Config} & \multicolumn{2}{c|}{CIFAR10} & \multicolumn{2}{c|}{CIFAR100} & \multicolumn{2}{c}{ ImageNet16-120} \\
 & Validation & Test & Validation & Test & Validation & Test \\
          \midrule
          w/o pruning-based sample & 90.83 & 93.39 & 71.08 & 70.98 & 44.83 & 43.98 \\        
          w/o objective reframing & 89.10 & 92.16 & 69.33 & 69.51 & 43.38 & 42.53 \\
          LLM-based Searcher (LLaMA-3.1-8B) & 91.52 & 94.20 & 72.82 & 72.96 & 46.48 & 46.51 \\
          LLM-based Searcher (Qwen3-8B) & 91.45 & 94.23 & 73.05 & 73.13 & 46.50 & 46.37 \\
          \bottomrule
        \end{tabular}
    }
    \label{tab:nas201_ablation}
\end{table*}

\subsection{Ablation Studies}\label{sec:ablations}
To validate the effectiveness of our approach and the resulting configurations, we compared our method against several baselines. As shown in Table~\ref{tab:geneval_ablation}, our joint architecture and hyperparameter search approach outperforms both architecture-only search and hyperparameter/data-ratio-only search by overall scores of 0.05 and 0.02, respectively. This highlights the critical importance of jointly optimizing model architectures, training hyperparameters, and data ratios. Furthermore, when applying the searched AutoML configurations without the searched architecture or without the hyperparameter and data configurations (reverting to manually designed baselines), the overall score drops by 0.07 and 0.09, respectively. These results demonstrate that our LLM-based searcher successfully navigates the optimal combination of all configurations, rather than treating their optimization as isolated problems.

Furthermore, to investigate the efficacy of the key components within our framework and evaluate the impact of different LLM backbones, we conduct a comprehensive ablation study on the NAS-Bench-201 benchmark (table~\ref{tab:nas201_ablation}). As delineated in Table~\ref{tab:nas201_ablation}, removing the pruning-based architecture sampling strategy (\textit{w/o} pruning-based sample) incurs a noticeable performance degradation across all three datasets (CIFAR-10, CIFAR-100, and ImageNet16-120), validating its critical role in augmenting trajectory diversity and structural distribution. More significantly, when disabling the objective reframing mechanism (\textit{w/o} objective reframing) and forcing the model to directly generate full NCodes rather than ranking candidates, we observe a catastrophic performance collapse (e.g., the test accuracy on ImageNet16-120 plummets from 46.51\% to 42.53\%). This sharp decline firmly demonstrates that reframing the discrete optimization into a tractable ranking task substantially alleviates the learning burden of the LLM searcher.

In addition, we compare the search efficiency of the LLM-based searchers built upon LLaMA-3.1-8B and Qwen3-8B. The empirical results indicate that both models achieve highly comparable and competitive performance across all validation and test sets (e.g., reaching 72.96\% and 73.13\% test accuracy on CIFAR-100, respectively). This comparable performance proves that instilling Chain-of-Thought (CoT) reasoning capabilities via Group Relative Policy Optimization (GRPO) does not jeopardize the core optimization efficacy. To optimize overall search efficiency and circumvent the substantial computational overhead associated with token-heavy CoT generation, we deploy the LLaMA-3.1-8B searcher to execute all primary search environments reported in this study. Concurrently, the Qwen3-8B searcher is leveraged exclusively for the interpretability analysis to unpack and systematically interpret the internal decision-making mechanisms and temporal strategic transitions of the universal search agent.

\subsection{Case Studies}\label{sec:case_study}
To further evaluate the fine-grained compositional text-to-image alignment of our optimized Diffusion Transformer, we present qualitative case studies comparing our method against the 12-billion-parameter FLUX.1-dev~\cite{flux2024} and the AutoML approach OptFormer~\cite{chen2022towards}. As illustrated in Fig.~\ref{fig:geneval_case_study}, our method demonstrates superior semantic understanding and instruction adherence across complex compositional prompts involving coloring, attribute binding, and spatial relationships

Specifically, in the first two columns evaluating color attribute binding, FLUX.1-dev struggles to overcome strong visual priors and assign atypical colors to objects. It generates a standard orange instead of the requested "red orange" and defaults to a standard red stop sign despite the prompt explicitly specifying a "white stop sign". In contrast, our method faithfully adheres to the instructions, rendering both the red orange and the white stop sign accurately. In the third column ("a photo of a purple wine glass and a black apple"), FLUX.1-dev misinterprets the attribute binding by applying the purple color to the liquid wine rather than the glass itself. Conversely, our model successfully disentangles these concepts to generate a purple-tinted glass.  

Furthermore, our method excels in understanding spatial dynamics and relational object associations. In the fourth column ("a photo of a vase right of a horse"), FLUX.1-dev fails to comprehend the distinct object relationship, suffering from concept bleeding where it erroneously generates a horse made of a porcelain material matching the vase. Similarly, in the fifth column ("a photo of a bird left of a couch"), FLUX.1-dev fails to reflect the correct relative spatial positioning requested by the user. Our model correctly captures these spatial and compositional dynamics in both instances. Finally, while OptFormer demonstrates a moderate degree of instruction adherence, it suffers from severe geometric distortions and a lack of photorealism. For instance, OptFormer generates a horse missing a leg (Column 4) and produces severely deformed, unrealistic couch structures (Column 5). These qualitative results underscore our LLM-based search framework's ability to discover model configurations that yield high-fidelity, photorealistic, and highly instruction-compliant visual generations.

\section{Conclusion and Future Work}
In this paper, we presented a universal LLM-based AutoML framework designed to jointly optimize model architectures, training hyperparameters, and data ratios across open-world scenarios without task-specific fine-tuning. By introducing NCode, a unified numerical encoding scheme, we mapped heterogeneous design choices into universal discrete sequences and reframed the optimization process as a conditional candidate ranking problem. Trained exclusively on trajectory data curated from a single-domain vision dataset, our searcher demonstrated remarkable zero-shot generalizability across seven domains, four tasks, and three modalities (vision, text, and audio). Experimental results validated its state-of-the-art efficiency and scalability, notably optimizing a 1.3-billion-parameter Diffusion Transformer to attain a GenEval score of 0.91. 

Furthermore, by instilling Chain-of-Thought reasoning through Group Relative Policy Optimization (GRPO), we demystified the searcher's internal decision-making mechanisms, revealing an adaptive transition from early-stage diversity-driven exploration to late-stage anchor exploitation. This work establishes a scalable, generalizable, and interpretable pathway toward autonomous, general-purpose machine learning optimization. Future work will explore extending the searcher to Vision-Language Models (VLMs) to integrate direct visual rewards for richer generative quality assessment.

\bibliography{reference}

@article{zela2018towards,
  title={Towards automated deep learning: Efficient joint neural architecture and hyperparameter search},
  author={Zela, Arber and Klein, Aaron and Falkner, Stefan and Hutter, Frank},
  journal={arXiv preprint arXiv:1807.06906},
  year={2018}
}

@inproceedings{nasir2024llmatic,
  title={Llmatic: neural architecture search via large language models and quality diversity optimization},
  author={Nasir, Muhammad Umair and Earle, Sam and Togelius, Julian and James, Steven and Cleghorn, Christopher},
  booktitle={proceedings of the Genetic and Evolutionary Computation Conference},
  pages={1110--1118},
  year={2024}
}

@inproceedings{hu2025lm,
  title={LM-Searcher: Cross-domain Neural Architecture Search with LLMs via Unified Numerical Encoding},
  author={Hu, Yuxuan and Liu, Jihao and Wang, Ke and Zheng, Jinliang and Shi, Weikang and Zhang, Manyuan and Dou, Qi and Liu, Rui and Zhou, Aojun and Li, Hongsheng},
  booktitle={Proceedings of the 2025 Conference on Empirical Methods in Natural Language Processing},
  pages={9419--9432},
  year={2025}
}

@String(ICASSP=	{ICASSP})

@inproceedings{rombach2022high,
  title={High-resolution image synthesis with latent diffusion models},
  author={Rombach, Robin and Blattmann, Andreas and Lorenz, Dominik and Esser, Patrick and Ommer, Bj{\"o}rn},
  booktitle={Proceedings of the IEEE/CVF conference on computer vision and pattern recognition},
  pages={10684--10695},
  year={2022}
}

@article{esser2024scaling,
  title={Scaling rectified flow transformers for high-resolution image synthesis},
  author={Esser, Patrick and Kulal, Sumith and Blattmann, Andreas and Entezari, Rahim and M{\"u}ller, Jonas and Saini, Harry and Levi, Yam and Lorenz, Dominik and Sauer, Axel and Boesel, Frederic and others},
  journal={arXiv preprint arXiv:2403.03206},
  year={2024}
}

@misc{openai2025gpt-image-1,
    author={OpenAI},
    title={GPT Image 1},
    year={2025},
    howpublished={\url{https://developers.openai.com/api/docs/models/gpt-image-1}},
}

@article{chen2022towards,
  title={Towards learning universal hyperparameter optimizers with transformers},
  author={Chen, Yutian and Song, Xingyou and Lee, Chansoo and Wang, Zi and Zhang, Richard and Dohan, David and Kawakami, Kazuya and Kochanski, Greg and Doucet, Arnaud and Ranzato, Marc'aurelio and others},
  journal={Advances in Neural Information Processing Systems},
  volume={35},
  pages={32053--32068},
  year={2022}
}

@inproceedings{zoph2017neural,
  title={Neural Architecture Search with Reinforcement Learning},
  author={Zoph, Barret and Le, Quoc},
  booktitle={International Conference on Learning Representations},
  year={2017}
}

@inproceedings{zoph2018learning,
  title={Learning transferable architectures for scalable image recognition},
  author={Zoph, Barret and Vasudevan, Vijay and Shlens, Jonathon and Le, Quoc V},
  booktitle={Proceedings of the IEEE conference on computer vision and pattern recognition},
  pages={8697--8710},
  year={2018}
}

@article{wei2022chain,
  title={Chain-of-thought prompting elicits reasoning in large language models},
  author={Wei, Jason and Wang, Xuezhi and Schuurmans, Dale and Bosma, Maarten and Xia, Fei and Chi, Ed and Le, Quoc V and Zhou, Denny and others},
  journal={Advances in neural information processing systems},
  volume={35},
  pages={24824--24837},
  year={2022}
}

@article{zheng2023can,
  title={Can gpt-4 perform neural architecture search?},
  author={Zheng, Mingkai and Su, Xiu and You, Shan and Wang, Fei and Qian, Chen and Xu, Chang and Albanie, Samuel},
  journal={arXiv preprint arXiv:2304.10970},
  year={2023}
}

@article{liu2018darts,
  title={Darts: Differentiable architecture search},
  author={Liu, Hanxiao and Simonyan, Karen and Yang, Yiming},
  journal={arXiv preprint arXiv:1806.09055},
  year={2018}
}

@inproceedings{huang2021searching,
  title={Searching by generating: Flexible and efficient one-shot NAS with architecture generator},
  author={Huang, Sian-Yao and Chu, Wei-Ta},
  booktitle={Proceedings of the IEEE/CVF Conference on Computer Vision and Pattern Recognition},
  pages={983--992},
  year={2021}
}

@article{asthana2024multi,
  title={Multi-conditioned Graph Diffusion for Neural Architecture Search},
  author={Asthana, Rohan and Conrad, Joschua and Dawoud, Youssef and Ortmanns, Maurits and Belagiannis, Vasileios},
  journal={arXiv preprint arXiv:2403.06020},
  year={2024}
}

@article{llama3modelcard,
  title={Llama 3 Model Card},
  author={AI@Meta},
  year={2024},
  url = {https://github.com/meta-llama/llama3/blob/main/MODEL_CARD.md}
}

@article{dong2020bench,
  title={Nas-bench-201: Extending the scope of reproducible neural architecture search},
  author={Dong, Xuanyi and Yang, Yi},
  journal={arXiv preprint arXiv:2001.00326},
  year={2020}
}

@article{xu2019pc,
  title={Pc-darts: Partial channel connections for memory-efficient architecture search},
  author={Xu, Yuhui and Xie, Lingxi and Zhang, Xiaopeng and Chen, Xin and Qi, Guo-Jun and Tian, Qi and Xiong, Hongkai},
  journal={arXiv preprint arXiv:1907.05737},
  year={2019}
}

@article{zela2020surrogate,
  title={Surrogate NAS benchmarks: Going beyond the limited search spaces of tabular NAS benchmarks},
  author={Zela, Arber and Siems, Julien and Zimmer, Lucas and Lukasik, Jovita and Keuper, Margret and Hutter, Frank},
  journal={arXiv preprint arXiv:2008.09777},
  year={2020}
}

@article{hu2021lora,
  title={Lora: Low-rank adaptation of large language models},
  author={Hu, Edward J and Shen, Yelong and Wallis, Phillip and Allen-Zhu, Zeyuan and Li, Yuanzhi and Wang, Shean and Wang, Lu and Chen, Weizhu},
  journal={arXiv preprint arXiv:2106.09685},
  year={2021}
}

@inproceedings{li2020random,
  title={Random search and reproducibility for neural architecture search},
  author={Li, Liam and Talwalkar, Ameet},
  booktitle={Uncertainty in artificial intelligence},
  pages={367--377},
  year={2020},
  organization={PMLR}
}

@inproceedings{kirillov2023segment,
  title={Segment anything},
  author={Kirillov, Alexander and Mintun, Eric and Ravi, Nikhila and Mao, Hanzi and Rolland, Chloe and Gustafson, Laura and Xiao, Tete and Whitehead, Spencer and Berg, Alexander C and Lo, Wan-Yen and others},
  booktitle={Proceedings of the IEEE/CVF international conference on computer vision},
  pages={4015--4026},
  year={2023}
}

@inproceedings{chen2023sam,
  title={Sam-adapter: Adapting segment anything in underperformed scenes},
  author={Chen, Tianrun and Zhu, Lanyun and Deng, Chaotao and Cao, Runlong and Wang, Yan and Zhang, Shangzhan and Li, Zejian and Sun, Lingyun and Zang, Ying and Mao, Papa},
  booktitle={Proceedings of the IEEE/CVF International Conference on Computer Vision},
  pages={3367--3375},
  year={2023}
}

@inproceedings{zhong2024convolution,
  title={Convolution meets lora: Parameter efficient finetuning for segment anything model},
  author={Zhong, Zihan and Tang, Zhiqiang and He, Tong and Fang, Haoyang and Yuan, Chun},
  booktitle={International Conference on Learning Representations},
  volume={2024},
  pages={26755--26779},
  year={2024}
}

@article{zaken2021bitfit,
  title={Bitfit: Simple parameter-efficient fine-tuning for transformer-based masked language-models},
  author={Zaken, Elad Ben and Ravfogel, Shauli and Goldberg, Yoav},
  journal={arXiv preprint arXiv:2106.10199},
  year={2021}
}

@inproceedings{houlsby2019parameter,
  title={Parameter-efficient transfer learning for NLP},
  author={Houlsby, Neil and Giurgiu, Andrei and Jastrzebski, Stanislaw and Morrone, Bruna and De Laroussilhe, Quentin and Gesmundo, Andrea and Attariyan, Mona and Gelly, Sylvain},
  booktitle={International conference on machine learning},
  pages={2790--2799},
  year={2019},
  organization={PMLR}
}

@article{sung2022lst,
  title={Lst: Ladder side-tuning for parameter and memory efficient transfer learning},
  author={Sung, Yi-Lin and Cho, Jaemin and Bansal, Mohit},
  journal={Advances in Neural Information Processing Systems},
  volume={35},
  pages={12991--13005},
  year={2022}
}

@article{lian2022scaling,
  title={Scaling \& shifting your features: A new baseline for efficient model tuning},
  author={Lian, Dongze and Zhou, Daquan and Feng, Jiashi and Wang, Xinchao},
  journal={Advances in Neural Information Processing Systems},
  volume={35},
  pages={109--123},
  year={2022}
}

@article{Chen2025BLIP3oAF,
  title={BLIP3-o: A Family of Fully Open Unified Multimodal Models-Architecture, Training and Dataset},
  author={Jiuhai Chen and Zhiyang Xu and Xichen Pan and Yushi Hu and Can Qin and Tom Goldstein and Lifu Huang and Tianyi Zhou and Saining Xie and Silvio Savarese and Le Xue and Caiming Xiong and Ran Xu},
  journal={ArXiv},
  year={2025},
  volume={abs/2505.09568},
  url={https://api.semanticscholar.org/CorpusID:278602689}
}

@article{Schuhmann2022LAION5BAO,
  title={LAION-5B: An open large-scale dataset for training next generation image-text models},
  author={Christoph Schuhmann and Romain Beaumont and Richard Vencu and Cade Gordon and Ross Wightman and Mehdi Cherti and Theo Coombes and Aarush Katta and Clayton Mullis and Mitchell Wortsman and Patrick Schramowski and Srivatsa Kundurthy and Katherine Crowson and Ludwig Schmidt and Robert Kaczmarczyk and Jenia Jitsev},
  journal={ArXiv},
  year={2022},
  volume={abs/2210.08402},
  url={https://api.semanticscholar.org/CorpusID:252917726}
}

@article{Wu2025OmniGen2ET,
  title={OmniGen2: Exploration to Advanced Multimodal Generation},
  author={Chenyuan Wu and Pengfei Zheng and Ruiran Yan and Shitao Xiao and Xin Luo and Yueze Wang and Wanli Li and Xiyan Jiang and Yexin Liu and Junjie Zhou and Ze Liu and Ziyi Xia and Chaofan Li and Haoge Deng and Jiahao Wang and Kun Luo and Bo Zhang and Defu Lian and Xinlong Wang and Zhongyuan Wang and Tiejun Huang and Zheng Liu},
  journal={ArXiv},
  year={2025},
  volume={abs/2506.18871},
  url={https://api.semanticscholar.org/CorpusID:279999713}
}

@article{Ye2025Echo4oHT,
  title={Echo-4o: Harnessing the Power of GPT-4o Synthetic Images for Improved Image Generation},
  author={Junyan Ye and Dongzhi Jiang and Zihao Wang and Leqi Zhu and Zhenghao Hu and Zilong Huang and Jun He and Zhiyuan Yan and Jing Yu and Hongsheng Li and Conghui He and Weijia Li},
  journal={ArXiv},
  year={2025},
  volume={abs/2508.09987},
  url={https://api.semanticscholar.org/CorpusID:280642445}
}

@article{Ghosh2023GenEvalAO,
  title={GenEval: An Object-Focused Framework for Evaluating Text-to-Image Alignment},
  author={Dhruba Ghosh and Hanna Hajishirzi and Ludwig Schmidt},
  journal={ArXiv},
  year={2023},
  volume={abs/2310.11513},
  url={https://api.semanticscholar.org/CorpusID:264288728}
}

@inproceedings{Yang2025Qwen3TR,
  title={Qwen3 Technical Report},
  author={An Yang and Anfeng Li and Baosong Yang and Beichen Zhang and Binyuan Hui and Bo Zheng and Bowen Yu and Chang Gao and Chengen Huang and Chenxu Lv and Chujie Zheng and Dayiheng Liu and Fan Zhou and Fei Huang and Feng Hu and Hao Ge and Haoran Wei and Huan Lin and Jialong Tang and Jian Yang and Jianhong Tu and Jianwei Zhang and Jianxin Yang and Jiaxin Yang and Jingren Zhou and Jingren Zhou and Junyan Lin and Kai Dang and Keqin Bao and Ke‐Pei Yang and Le Yu and Li-Chun Deng and Mei Li and Min Xue and Mingze Li and Pei Zhang and Peng Wang and Qin Zhu and Rui Men and Ruize Gao and Shi-Qiang Liu and Shuang Luo and Tianhao Li and Tianyi Tang and Wenbiao Yin and Xingzhang Ren and Xinyu Wang and Xinyu Zhang and Xuancheng Ren and Yang Fan and Yang Su and Yi-Chao Zhang and Yinger Zhang and Yu Wan and Yuqiong Liu and Zekun Wang and Zeyu Cui and Zhenru Zhang and Zhipeng Zhou and Zihan Qiu},
  year={2025},
  url={https://api.semanticscholar.org/CorpusID:278602855}
}

@article{shao2024deepseekmath,
  title={Deepseekmath: Pushing the limits of mathematical reasoning in open language models},
  author={Shao, Zhihong and Wang, Peiyi and Zhu, Qihao and Xu, Runxin and Song, Junxiao and Bi, Xiao and Zhang, Haowei and Zhang, Mingchuan and Li, YK and Wu, Yang and others},
  journal={arXiv preprint arXiv:2402.03300},
  year={2024}
}

@inproceedings{xiesnas,
  title={SNAS: stochastic neural architecture search},
  author={Xie, Sirui and Zheng, Hehui and Liu, Chunxiao and Lin, Liang},
  booktitle={International Conference on Learning Representations},
  year={2019}
}

@inproceedings{chen20253d,
  title={3D-speaker-toolkit: An open-source toolkit for multimodal speaker verification and diarization},
  author={Chen, Yafeng and Zheng, Siqi and Wang, Hui and Cheng, Luyao and Zhu, Tinglong and Huang, Rongjie and Deng, Chong and Chen, Qian and Zhang, Shiliang and Wang, Wen and others},
  booktitle={ICASSP 2025-2025 IEEE International Conference on Acoustics, Speech and Signal Processing (ICASSP)},
  pages={1--5},
  year={2025},
  organization={IEEE}
}

@article{Fan2017StructureMeasureAN,
  title={Structure-Measure: A New Way to Evaluate Foreground Maps},
  author={Deng-Ping Fan and Ming-Ming Cheng and Yun Liu and Tao Li and Ali Borji},
  journal={International Journal of Computer Vision},
  year={2017},
  volume={129},
  pages={2622 - 2638},
  url={https://api.semanticscholar.org/CorpusID:22726592}
}

@inproceedings{Fan2018EnhancedalignmentMF,
  title={Enhanced-alignment Measure for Binary Foreground Map Evaluation},
  author={Deng-Ping Fan and Cheng Gong and Yang Cao and Bo Ren and Ming-Ming Cheng and Ali Borji},
  booktitle={International Joint Conference on Artificial Intelligence},
  year={2018},
  url={https://api.semanticscholar.org/CorpusID:44072899}
}

@inproceedings{jia2022visual,
  title={Visual prompt tuning},
  author={Jia, Menglin and Tang, Luming and Chen, Bor-Chun and Cardie, Claire and Belongie, Serge and Hariharan, Bharath and Lim, Ser-Nam},
  booktitle={European conference on computer vision},
  pages={709--727},
  year={2022},
  organization={Springer}
}

@article{gao2019res2net,
  title={Res2net: A new multi-scale backbone architecture},
  author={Gao, Shang-Hua and Cheng, Ming-Ming and Zhao, Kai and Zhang, Xin-Yu and Yang, Ming-Hsuan and Torr, Philip},
  journal={IEEE transactions on pattern analysis and machine intelligence},
  volume={43},
  number={2},
  pages={652--662},
  year={2019},
  publisher={IEEE}
}

@incollection{koonce2021resnet,
  title={{ResNet} 34},
  author={Koonce, Brett},
  booktitle={Convolutional Neural Networks with Swift for {TensorFlow}: Image Recognition and Dataset Categorization},
  pages={51--61},
  year={2021},
  publisher={Apress},
  address={Berkeley, CA},
  doi={10.1007/978-1-4842-6168-2_5}
}

@article{desplanques2020ecapa,
  title={Ecapa-tdnn: Emphasized channel attention, propagation and aggregation in tdnn based speaker verification},
  author={Desplanques, Brecht and Thienpondt, Jenthe and Demuynck, Kris},
  journal={arXiv preprint arXiv:2005.07143},
  year={2020}
}

@article{chen2023enhanced,
  title={An enhanced res2net with local and global feature fusion for speaker verification},
  author={Chen, Yafeng and Zheng, Siqi and Wang, Hui and Cheng, Luyao and Chen, Qian and Qi, Jiajun},
  journal={arXiv preprint arXiv:2305.12838},
  year={2023}
}

@article{wang2023cam++,
  title={Cam++: A fast and efficient network for speaker verification using context-aware masking},
  author={Wang, Hui and Zheng, Siqi and Chen, Yafeng and Cheng, Luyao and Chen, Qian},
  journal={arXiv preprint arXiv:2303.00332},
  year={2023}
}

@inproceedings{podell2024sdxl,
  title={Sdxl: Improving latent diffusion models for high-resolution image synthesis},
  author={Podell, Dustin and English, Zion and Lacey, Kyle and Blattmann, Andreas and Dockhorn, Tim and M{\"u}ller, Jonas and Penna, Joe and Rombach, Robin},
  booktitle={International Conference on Learning Representations},
  volume={2024},
  pages={1862--1874},
  year={2024}
}

@misc{flux2024,
    author={Black Forest Labs},
    title={FLUX},
    year={2024},
    howpublished={\url{https://github.com/black-forest-labs/flux}},
}

@article{wang2024emu3,
  title={Emu3: Next-token prediction is all you need},
  author={Wang, Xinlong and Zhang, Xiaosong and Luo, Zhengxiong and Sun, Quan and Cui, Yufeng and Wang, Jinsheng and Zhang, Fan and Wang, Yueze and Li, Zhen and Yu, Qiying and others},
  journal={arXiv preprint arXiv:2409.18869},
  year={2024}
}

@inproceedings{ma2025janusflow,
  title={Janusflow: Harmonizing autoregression and rectified flow for unified multimodal understanding and generation},
  author={Ma, Yiyang and Liu, Xingchao and Chen, Xiaokang and Liu, Wen and Wu, Chengyue and Wu, Zhiyu and Pan, Zizheng and Xie, Zhenda and Zhang, Haowei and Yu, Xingkai and others},
  booktitle={Proceedings of the IEEE/CVF Conference on Computer Vision and Pattern Recognition},
  pages={7739--7751},
  year={2025}
}

@article{chen2025janus,
  title={Janus-pro: Unified multimodal understanding and generation with data and model scaling},
  author={Chen, Xiaokang and Wu, Zhiyu and Liu, Xingchao and Pan, Zizheng and Liu, Wen and Xie, Zhenda and Yu, Xingkai and Ruan, Chong},
  journal={arXiv preprint arXiv:2501.17811},
  year={2025}
}

@inproceedings{xiao2025omnigen,
  title={Omnigen: Unified image generation},
  author={Xiao, Shitao and Wang, Yueze and Zhou, Junjie and Yuan, Huaying and Xing, Xingrun and Yan, Ruiran and Li, Chaofan and Wang, Shuting and Huang, Tiejun and Liu, Zheng},
  booktitle={Proceedings of the IEEE/CVF Conference on Computer Vision and Pattern Recognition},
  pages={13294--13304},
  year={2025}
}

@article{deng2025emerging,
  title={Emerging properties in unified multimodal pretraining},
  author={Deng, Chaorui and Zhu, Deyao and Li, Kunchang and Gou, Chenhui and Li, Feng and Wang, Zeyu and Zhong, Shu and Yu, Weihao and Nie, Xiaonan and Song, Ziang and others},
  journal={arXiv preprint arXiv:2505.14683},
  year={2025}
}

@inproceedings{qin2025lumina,
  title={Lumina-image 2.0: A unified and efficient image generative framework},
  author={Qin, Qi and Zhuo, Le and Xin, Yi and Du, Ruoyi and Li, Zhen and Fu, Bin and Lu, Yiting and Li, Xinyue and Liu, Dongyang and Zhu, Xiangyang and others},
  booktitle={Proceedings of the IEEE/CVF International Conference on Computer Vision},
  pages={20031--20042},
  year={2025}
}

@inproceedings{cai2025hidream,
  title={HiDream-I1: An Open-Source High-Efficient Image Generative Foundation Model},
  author={Cai, Qi and Li, Yehao and Pan, Yingwei and Yao, Ting and Mei, Tao},
  booktitle={Proceedings of the 33rd ACM International Conference on Multimedia},
  pages={13636--13639},
  year={2025}
}

@article{wu2025qwen,
  title={Qwen-image technical report},
  author={Wu, Chenfei and Li, Jiahao and Zhou, Jingren and Lin, Junyang and Gao, Kaiyuan and Yan, Kun and Yin, Sheng-ming and Bai, Shuai and Xu, Xiao and Chen, Yilei and others},
  journal={arXiv preprint arXiv:2508.02324},
  year={2025}
}

@article{gao2025seedream,
  title={Seedream 3.0 technical report},
  author={Gao, Yu and Gong, Lixue and Guo, Qiushan and Hou, Xiaoxia and Lai, Zhichao and Li, Fanshi and Li, Liang and Lian, Xiaochen and Liao, Chao and Liu, Liyang and others},
  journal={arXiv preprint arXiv:2504.11346},
  year={2025}
}

@article{zhang2023automl,
  title={Automl-gpt: Automatic machine learning with gpt},
  author={Zhang, Shujian and Gong, Chengyue and Wu, Lemeng and Liu, Xingchao and Zhou, Mingyuan},
  journal={arXiv preprint arXiv:2305.02499},
  year={2023}
}

@inproceedings{li2026collm,
  title={CoLLM-NAS: Collaborative large language models for efficient knowledge-guided neural architecture search},
  author={Li, Zhe and Lin, Zhiwei and Wang, Yongtao},
  booktitle={Proceedings of the IEEE/CVF Conference on Computer Vision and Pattern Recognition},
  pages={3273--3282},
  year={2026}
}

@inproceedings{hutter2014efficient,
  title={An efficient approach for assessing hyperparameter importance},
  author={Hutter, Frank and Hoos, Holger and Leyton-Brown, Kevin},
  booktitle={International conference on machine learning},
  pages={754--762},
  year={2014},
  organization={PMLR}
}
\bibliographystyle{IEEEtran}

\vfill

\end{document}